\documentclass[smallextended,referee,envcountsect]{svjour3} 
\smartqed 
\usepackage{amsmath}
\usepackage{hyperref}       
\usepackage{url}            
\usepackage{booktabs}       
\usepackage{amsfonts}       
\usepackage{nicefrac}       
\usepackage{microtype}      
\usepackage{graphicx} 
\usepackage{amssymb}
\usepackage{array}
\usepackage{multirow}
\usepackage{subcaption}
\usepackage{float}
\usepackage[mathscr]{euscript}
\usepackage{algorithm}
\usepackage{algpseudocode}
\usepackage{xcolor}
\usepackage{color}
\usepackage{booktabs}
\usepackage{pgfplots}
\usepackage{pgfplotstable}
\usepackage{array}
\usepackage{hhline}
\usepackage{mathtools}

\renewcommand\refname{\vspace{-.3in}}
\newcommand{\vertiii}[1]{{\left\vert\kern-0.25ex\left\vert\kern-0.25ex\left\vert #1 
		\right\vert\kern-0.25ex\right\vert\kern-0.25ex\right\vert}}

\newcommand{\benon}{\begin{equation*}}  
\newcommand{\bemuln}[1]{\begin{multline}\label{#1}}
\newcommand{\bemul}{\begin{multline*}}
\newcommand{\been}[1]{\begin{eqnarray}\label{#1}}
\newcommand{\eeen}{\end{eqnarray}}
\newcommand{\began}[1]{\begin{gather}\label{#1}}
\newcommand{\bega}{\begin{gather*}}
\newcommand{\bealn}[1]{\begin{align}\label{#1}}
\newcommand{\beal}{\begin{align*}}
\newcommand{\bealatn}[2]{\begin{alignat}{#1}\label{#2}}
\newcommand{\bealat}{\begin{alignat*}}
\newcommand{\bexalatn}[1]{\begin{xalignat}\label{#1}}
\newcommand{\bexalat}{\begin{xalignat*}}

\newcommand{\mbb}{\mathbb}

\newtheorem{thm}{Theorem}[section]

\newtheorem{lem}[thm]{Lemma}
\newtheorem{col}[thm]{Corollary}
\newtheorem{defi}{Definition}
\newtheorem{ass}{Assumption} 
 
\def\ba{{\mathbf a}}
\def\bb{{\mathbf b}}

\def\bg{{\mathbf g}}

\def\bu{{\mathbf u}}
\def\bv{{\mathbf v}}
\def\bw{{\mathbf w}}
\def\bx{{\mathbf x}}  
\def\by{{\mathbf y}}
\def\bz{{\mathbf z}}

\def\bI{{\mathbf I}}

\def\bX{{\mathbf X}}

\def\bZ{{\mathbf Z}}

\def\texitem#1{\par\smallskip\noindent\hangindent 25pt
               \hbox to 25pt {\hss #1 ~}\ignorespaces}

\newcommand{\bzero}{{\mathbf{0}}}

\newcommand{\scrA}{\mathcal{A}}
\newcommand{\scrB}{\mathcal{B}}

\newcommand{\scrG}{\mathcal{G}}

\newcommand{\scrL}{\mathcal{L}}

\newcommand{\scrN}{\mathcal{N}}

\newcommand{\scrP}{\mathcal{P}}

\newcommand{\scrR}{\mathcal{R}}

\newcommand{\scrZ}{\mathcal{Z}}

\newcommand{\bbeta}{\boldsymbol{\beta}}
\newcommand{\bGamma}{\boldsymbol{\Gamma}}

\newcommand{\bEta}{\boldsymbol{\eta}}
\newcommand{\btheta}{\boldsymbol{\theta}}

\newcommand{\bSigma}{{\boldsymbol{\Sigma}}}

\journalname{**}

\begin{document}

\title{Robust Grouped Variable Selection Using Distributionally Robust Optimization}


\author{Ruidi Chen   \and  Ioannis Ch. Paschalidis }

\institute{Ruidi Chen \at
             Boston University \\
              Boston, MA, USA\\
              rchen15@bu.edu 
           \and
           Ioannis Ch. Paschalidis,  Corresponding author  \at
              Boston University \\
              Boston, MA, USA\\
              yannisp@bu.edu
}

\date{}

\maketitle

\begin{abstract}
We propose a {\em Distributionally Robust Optimization (DRO)} formulation with a
Wasserstein-based uncertainty set for selecting grouped variables under perturbations
on the data for both linear regression and classification problems. The resulting
model offers robustness explanations for {\em Grouped Least Absolute Shrinkage and
	Selection Operator (GLASSO)} algorithms and highlights the connection between
robustness and regularization. We prove probabilistic bounds on the out-of-sample
loss and the estimation bias, and establish the grouping effect of our estimator,
showing that coefficients in the same group converge to the same value as the sample
correlation between covariates approaches 1. Based on this result, we propose to use
the spectral clustering algorithm with the Gaussian similarity function to perform
grouping on the predictors, which makes our approach applicable without knowing the
grouping structure a priori. We compare our approach to an array of alternatives and
provide extensive numerical results on both synthetic data and a real large dataset
of surgery-related medical records, showing that our formulation produces an
interpretable and parsimonious model that encourages sparsity at a group level and is
able to achieve better prediction and estimation performance in the presence of
outliers.
\end{abstract}


\section{Introduction} \label{sec:intro}

We consider the problem of finding a robust regression/classification plane under
perturbations on the training data, when there exists a predefined grouping structure
for the predictors, e.g., encoding a categorical predictor using a group of indicator
variables. The goal is to jointly select/drop all variables in a group, i.e., induce
{\em group sparsity}, and produce robust estimates that generalize well out of
sample. Grouped variable selection gives rise to more interpretable models. Moreover,
group sparsity leads to an estimation error of regression coefficients that scales
with the number of groups and group sizes, instead of with the raw number of features
in the regression model \cite{huang2010benefit,lounici2011oracle}.

To perform variable selection at a group level, the {\em Grouped Least Absolute Shrinkage and Selection Operator (GLASSO)} was proposed by \cite{bakin1999adaptive,yuan2006model}.
Several extensions have been explored in later works, see
\cite{zhao2009composite,jacob2009group,simon2013sparse,bunea2014group}. The
group sparsity in general regression/classification models has also been
investigated, see, for example, \cite{meier2008group} for GLASSO in logistic
regression, and \cite{roth2008group} for GLASSO in generalized linear models.  We
note that most of the existing works endeavor to generalize/modify the GLASSO
formulation heuristically to achieve various goals. However, few of those works were
able to provide a rigorous explanation or theoretical justification for the form of
the penalty term.

In this work we attempt to fill this gap by casting the robust grouped variable
selection problem into a {\em Distributionally Robust Optimization (DRO)}
framework,
which induces robustness via minimizing a worst-case expected loss function
over a probabilistic ambiguity set that is constructed from the observed samples and
characterized by certain known properties of the true data-generating
distribution. DRO has been an active area of research in recent years, due to its
probabilistic interpretation of the uncertain data, tractability when assembled with
certain metrics, and extraordinary performance observed on numerical examples, see,
for example, \cite{gao2016distributionally,gao2017wasserstein,shafieezadeh2017regularization,Pey15,chen2017outlier}.  The uncertainty set in
DRO can be constructed (i) through a moment ambiguity set \cite{Ye10,goh2010distributionally,zymler2013distributionally}, or (ii) as a ball of
distributions centered at some nominal distribution defined via some probabilistic
distance metric such as the $\phi$-divergence, the Prokhorov metric, and the
Wasserstein distance.

We consider a DRO formulation with the uncertainty set being a ball of distributions
defined via the Wasserstein metric, motivated by the fact that (i) the Wasserstein
metric takes into account the closeness between support points while other metrics
only consider the probabilities on these points, and (ii) the Wasserstein ambiguity
set is rich enough to contain both continuous and discrete relevant distributions,
while other metrics such as the Kullback-Leibler (KL) divergence, do not allow for
probability mass outside the support of the nominal distribution.  We show that in
{\em Least Absolute Deviation (LAD)} and {\em logistic regression (LG)}, for both
non-overlapping and overlapping groups, by using a specific norm-induced Wasserstein
metric, the Wasserstein DRO model can be reformulated as a regularized empirical loss
minimization problem, where the regularizer coincides with the GLASSO penalty, and
its magnitude is equal to the radius of the distributional ambiguity set. Through
such a reformulation we establish a connection between regularization and robustness
and offer new insights into the GLASSO penalty term.

We should note that such a connection between robustification and regularization
under norm-bounded deterministic disturbances in the predictors has been discovered
in \cite{xu2009robust,yang2013unified,bertsimas2017characterization}. Within the
Wasserstein DRO framework, such an equivalence has been established for LG in
\cite{shafieezadeh2015distributionally}, and for LAD regression in
\cite{chen2017outlier}. More recently, \cite{shafieezadeh2017regularization,gao2017wasserstein} have provided a unified framework for connecting the
Wasserstein DRO with regularized learning procedures. None of the aforementioned
works, however, considered grouped variable selection; our work sheds new light on
the significance of exploring the group-wise DRO problem. It is worth noting that
\cite{blanchet2017distributionally} has studied the group-wise regularization
estimator with the square root of the expected loss under the Wasserstein DRO
framework and recovered the {\em Grouped Square Root LASSO (GSRL)}. Here, we present
a more general framework that includes both the LAD and the negative log-likelihood
loss functions, under both non-overlapping and overlapping group
structures. Moreover, we point out the potential of generalizing such results to a
class of loss functions with a finite growth rate.

Another contribution of this work lies in adding a correlation-based pre-clustering
step to GLASSO, as a consequence of a grouping effect result derived specifically for
our DRO GLASSO estimator. This has a similar flavor to \cite{buhlmann2013correlated},
where they considered a pre-clustering step based on either the canonical correlation
between groups or the sample correlation between covariates and validated their
approach from the standpoint of statistical consistency. Here, we justify the
correlation-based clustering from the optimization point of view, by analyzing the
optimality conditions satisfied by the DRO GLASSO estimator.

The remainder of the paper is organized as follows. Section~\ref{s2} introduces the
Wasserstein GLASSO formulations for LAD and LG. Section~\ref{s3} establishes a
desirable grouping effect for the solutions, which leads to a correlation-based pre-clustering step on
the predictors. 
Section~\ref{s4} presents numerical results on both synthetic
data and a real very large dataset with surgery-related medical records.
Conclusions are in Section~\ref{s5}. 

\textbf{Notational conventions:} We use boldfaced lowercase letters to denote
vectors, ordinary lowercase letters to denote scalars, boldfaced uppercase letters to
denote matrices, and calligraphic capital letters to denote sets. $\mathbb{E}$
denotes expectation and $\mathbb{P}$ probability of an event. All vectors are column
vectors. For space saving reasons, we write $\bx=(x_1, \ldots, x_n)$ to denote the
column vector $\bx \in \mbb{R}^n$. We use prime to denote transpose, $\|\cdot\|$ for
the general norm operator, and $\|\bx\|_p\triangleq(\sum_i |x_i|^p)^{1/p}$ for the
$\ell_p$ norm, where $p\geq 1$.

\section{Problem Formulation} \label{s2} 

In this section we describe the model setup and derive what we call the {\em
	Groupwise Wasserstein Grouped LASSO (GWGL)} formulation for  an LAD regression model and an LG
model. 

\subsection{GWGL for Continuous Response Variables}
\label{sec:gwgl_lr} 
Consider a linear regression model:
\begin{equation} \label{true}
\by = \bX \bbeta^* + \bEta, 
\end{equation}
where $\by = (y_1, \ldots, y_N)$ is the response vector, $\bX$ is an $N \times p$
design matrix, with $i$-th row $\bx_i'$ being the predictor vector for the $i$-th
sample, $\bbeta^* \in \mbb{R}^p$ is the vector of regression coefficients, and $\bEta
\in \mbb{R}^N$ is a random noise vector. We assume that the predictors belong to $L$
prescribed groups, with group size $p_l$, $l=1, \ldots, L$, and $\sum_{l=1}^L p_l =
p$ (no overlap among groups).  We use $\bx_{,j} \in \mbb{R}^N$ to denote the $j$-th
column of $\bX$, corresponding to the $j$-th predictor. A $p_l$-dimensional vector
$\bbeta^l$ denotes the vector of regression coefficients for group $l$. For a generic
predictor vector $\bx \in \mbb{R}^p$, we decompose it into $L$ groups $\bx = (\bx^1,
\ldots, \bx^L)$, each $\bx^l$ containing the $p_l$ predictors of group $l$.

The main assumption we make regarding $\bbeta^*$ is that it is {\em group sparse},
i.e., $\bbeta^l = \bzero$ for $l$ in some subset of $\{1, \ldots, L\}$. 
Our goal is
to obtain an accurate estimate of $\bbeta^*$ under perturbations on $(\bX, \by)$. Suppose we have $N$ i.i.d. samples $(\bx_i, y_i)$,
$i=1,\ldots,N$. We
model stochastic disturbances on the data via distributional uncertainty, and apply a
Wasserstein DRO framework to inject robustness into the solution. Our learning
problem is formulated as:
\begin{equation} \label{dro}
\inf\limits_{\bbeta}\sup\limits_{\mbb{Q}\in \Omega}
\mbb{E}^{\mbb{Q}}\big[ |y-\bx'\bbeta|\big], 
\end{equation}
where $(\bx, y) \in \mbb{R}^{p+1}$ denotes a generic predictor-response pair; and
$\mbb{Q}$ is the probability distribution of $(\bx, y)$. The inner optimization
problem is over $\mbb{Q}$ in some set $\Omega$ defined as:
\begin{equation} \label{omega}
\Omega \triangleq \{\mbb{Q}\in \scrP(\scrZ): W_1(\mathbb{Q},\
\hat{\mathbb{P}}_N) \le \epsilon\}, 
\end{equation}
where $\epsilon$ is a non-negative quantity determining the size of the ambiguity set
$\Omega$, $\scrZ$ is the set of possible values for $(\bx, y)$, $\scrP(\scrZ)$ is the
space of all probability distributions supported on $\scrZ$, $\hat{\mbb{P}}_N$ is the
empirical probability distribution that assigns equal probability on each training
sample point $(\bx_i, y_i)$, $i=1,\ldots,N$, and $W_1(\mbb{Q},\ \hat{\mbb{P}}_N)$ is
the order-one Wasserstein distance between $\mbb{Q}$ and $\hat{\mbb{P}}_N$ defined on
the metric space $(\scrZ, s)$ by:
\begin{equation} \label{wass_p}
W_1 (\mbb{Q}, \ \hat{\mbb{P}}_N) \triangleq \min\limits_{\Pi \in \scrP(\scrZ \times \scrZ)} \biggl\{\int_{\scrZ \times \scrZ} s((\bx_1, y_1), (\bx_2, y_2)) \ \Pi \bigl(d(\bx_1, y_1), d(\bx_2, y_2)\bigr)\biggr\},
\end{equation}
where we use the metric $s((\bx_1, y_1), (\bx_2, y_2)) = \|(\bx_1, y_1) - (\bx_2,
y_2)\|$ for the regression setting; and $\Pi$ is the joint distribution of $(\bx_1,
y_1)$ and $(\bx_2, y_2)$ with marginals $\mbb{Q}$ and $\hat{\mbb{P}}_N$,
respectively.

We assume that all the $N$ training samples $(\bx_i,y_i), i = 1, \ldots, N$, are
independent and identical realizations of $(\bx, y)$, which comes from a mixture of
two distributions, with probability $q$ from an ``outlying'' distribution
$\mathbb{P}_{\text{out}}$ and with probability $1-q$ from the true distribution
$\mathbb{P}$. Our goal is to generate estimators that are consistent with the true
distribution $\mathbb{P}$. We next show that if $q<0.5$, and $\epsilon$ chosen
judiciously, this is possible.
\begin{thm} \label{mixture}
	Suppose we are given two probability distributions $\mathbb{P}$ and $\mathbb{P}_{\text{out}}$, and the mixture distribution $\mathbb{P}_{\text{mix}}$ is a convex combination of the two:
	$\mbb{P}_{\text{mix}} = q \mathbb{P}_{\text{out}} + (1-q)\mathbb{P}$. Then,
	\begin{equation*}
	\frac{W_1(\mathbb{P}_{\text{out}}, \mathbb{P}_{\text{mix}})}{W_1(\mathbb{P}, \mathbb{P}_{\text{mix}})} = \frac{1- q}{q}.
	\end{equation*}
\end{thm}

Theorem \ref{mixture} implies that when $q<0.5$, and $W_1(\mathbb{P}, \mathbb{P}_{\text{mix}}) \le \epsilon< W_1(\mathbb{P}_{\text{out}}, \mathbb{P}_{\text{mix}})$, for a large enough sample size (so that $\hat{\mbb{P}}_N$ is a good approximation of $\mathbb{P}_{\text{mix}}$), the probabilistic ambiguity set $\Omega$ will include the true distribution and exclude the outlying one, thus providing protection against the disturbances.

The formulation in (\ref{dro}) is robust since it minimizes over the regression
coefficients the worst case expected loss; the latter being the expected loss
maximized over all probability distributions in the ambiguity set
$\Omega$. Formulation (\ref{dro}) injects additional robustness by adopting the LAD
loss, rendering it more robust to large residuals and yielding a smaller estimation
bias \cite{chen2017outlier}.

It has been shown in \cite{chen2017outlier} that (\ref{dro}) could be relaxed to:
\begin{equation} \label{qcp}
\inf\limits_{\bbeta} \frac{1}{N}\sum\limits_{i=1}^N|y_i - \bx_i'\bbeta| + \epsilon\|(-\bbeta, 1)\|_*,
\end{equation}
where $\|\cdot\|_*$ is the dual norm of $\|\cdot\|$ defined as $\|\btheta\|_*
\triangleq \sup_{\|\bz\|\le 1}\btheta'\bz$. Our GWGL formulation will be derived as a
special case of (\ref{qcp}), using a specific notion of norm on the $(\bx, y)$ space
that reflects the group structure of the predictors and takes into account the group
sparsity requirement.  Specifically, for a vector $\bz$ with a group structure $\bz =
(\bz^1, \ldots, \bz^L)$, define its $(q, t)$-norm, with $q,t\geq 1$, as:
\begin{equation*}
\|\bz\|_{q, t} = \Bigl(\sum_{l=1}^L
\bigl(\|\bz^l\|_q\bigr)^t\Bigr)^{1/t}. 
\end{equation*}
The $(q, t)$-norm of $\bz$ is actually the $\ell_t$-norm of the vector $(\|\bz^1\|_q,
\ldots, \|\bz^L\|_q)$, which represents each group vector $\bz^l$ in a concise way
via the $\ell_q$-norm.

Inspired by the LASSO where the $\ell_1$-regularizer is used to induce sparsity on
the individual level, we wish to deduce an $\ell_1$-norm penalty from (\ref{qcp}) on
the group level to induce group sparsity on $\bbeta^*$. This motivates the use of the
$(2, \infty)$-norm on the weighted predictor-response vector $\bz_{\bw} \triangleq
(\frac{1}{\sqrt{p_1}}\bx^1, \ldots, \frac{1}{\sqrt{p_L}}\bx^L, M y)$, where the
weight vector is $\bw=(\frac{1}{\sqrt{p_1}}, \ldots, \frac{1}{\sqrt{p_L}}, M)$, and
$M$ is a positive weight assigned to the response. Specifically,
\begin{equation} \label{2infty}
\|\bz_{\bw}\|_{2, \infty} = \max \left\{\frac{1}{\sqrt{p_1}}\|\bx^1\|_2,
\ldots, \frac{1}{\sqrt{p_L}}\|\bx^L\|_2, M |y|\right\}.
\end{equation}

In (\ref{2infty}) we normalize each group by the number of predictors, to prevent
large groups from having a large impact on the distance metric. The $\|\cdot\|_{2,
	\infty}$ operator computes the maximum of the $\ell_2$ norms of the (weighted)
grouped predictors and the response. It essentially selects the most influential
group when determining the closeness between two points in the predictor-response
space, which is consistent with our group sparsity assumption in that not all groups
of predictors contribute to the determination of $y$, and thus a metric that ignores
the unimportant groups (e.g., $\|\cdot\|_{2, \infty}$) is desired.

To obtain the GWGL formulation, we need to derive the dual norm of $\|\cdot\|_{2,
	\infty}$. A general result that applies to any $(q, t)$-norm is presented in the
following theorem. 
\begin{thm} \label{dualnorm} Consider a vector $\bx = (\bx^1,
	\ldots, \bx^L)$, where each $\bx^l \in \mbb{R}^{p_l}$, and $\sum_l p_l
	= p$. Define the weighted $(r, s)$-norm of $\bx$ with the weight
	vector $\bw = (w_1, \ldots, w_L)$ to be:
	\begin{equation*}
	\|\bx_{\bw}\|_{r, s} = \Bigl(\sum_{l=1}^L
	\bigl(\|w_l\bx^l\|_r\bigr)^s\Bigr)^{1/s}, 
	\end{equation*} 
	where $\bx_{\bw} = (w_1\bx^1, \ldots, w_L\bx^L)$, $w_l>0, \forall
	l$, and $r, s \ge 1$. Then, the dual norm of the weighted $(r, s)$-norm with weight $\bw$
	is the $(q, t)$-norm with weight $\bw^{-1}$, where $1/r + 1/q = 1$,
	$1/s+1/t = 1$, and $\bw^{-1} = (1/w_1, \ldots, 1/w_L)$.
\end{thm} 

Now, let us go back to (\ref{2infty}), which is the weighted $(2, \infty)$-norm of
$\bz = (\bx^1, \ldots, \bx^L, y)$ with the weight $\bw = (\frac{1}{\sqrt{p_1}},
\ldots, \frac{1}{\sqrt{p_L}}, M)$.  According to Theorem \ref{dualnorm}, the dual
norm of the weighted $(2, \infty)$-norm with weight $\bw$ evaluated at some
$\tilde{\bbeta} = (-\bbeta^1, \ldots, -\bbeta^L, 1)$ is:
\begin{equation*}
\|\tilde{\bbeta}_{\bw^{-1}}\|_{2, 1} = \sum_{l=1}^L \sqrt{p_l}\|\bbeta^l\|_2 + \frac{1}{M},
\end{equation*} 
where $\bw^{-1} = (\sqrt{p_1}, \ldots, \sqrt{p_L}, 1/M)$. Therefore, the GWGL
formulation for Linear Regression (GWGL-LR) takes the following form:
\begin{equation} \label{gwgl-lr} 
\inf\limits_{\bbeta} 
\frac{1}{N}\sum\limits_{i=1}^N|y_i - \bx_i' \bbeta| + \epsilon  
\sum_{l=1}^L \sqrt{p_l}\|\bbeta^l\|_2,
\end{equation}
where the constant term $1/M$ has been removed. We see that by using the weighted
$(2, \infty)$-norm in the predictor-response space, we are able to recover the
commonly used penalty term for GLASSO \cite{bakin1999adaptive,yuan2006model}. Our
Wasserstein DRO framework offers new interpretations for the GLASSO penalty from the
standpoint of the distance metric on the predictor-response space and establishes the
connection between group sparsity and distributional robustness.

\subsection{GWGL for Binary Categorical Response Variables}
\label{sec:gwgl_lg}
In this subsection we will explore the GWGL formulation for binary classification
problems. Let $\bx \in \mbb{R}^p$ denote the predictor and $y \in \{-1, +1\}$ the
associated binary label to be predicted. In LG, the conditional distribution
of $y$ given $\bx$ is modeled as
\begin{equation*}
\mbb{P}(y|\bx) = \big(1+\exp(-y \bbeta'\bx)\big)^{-1},
\end{equation*}
where $\bbeta \in \mbb{R}^p$ is the unknown coefficient vector (classifier) to be
estimated. The {\em Maximum Likelihood Estimator (MLE)} of $\bbeta$ is found by
minimizing the {\em negative log-likelihood (logloss)}:  
\begin{equation*}
l_{\bbeta}(\bx, y) = \log(1+\exp(-y \bbeta'\bx)).
\end{equation*}
To apply the Wasserstein DRO framework, we define the distance metric on the
predictor-response space as follows.
\begin{equation} \label{metric-lg}
s((\bx_1, y_1), (\bx_2, y_2)) \triangleq \|\bx_1 - \bx_2\| + M |y_1 - y_2|, \ \forall (\bx_1, y_1), (\bx_2, y_2) \in \scrZ,
\end{equation}
where $M$ is an infinitely large positive number (different from Section
\ref{sec:gwgl_lr} where $M$ could be any positive number), and $\scrZ = \mbb{R}^p
\times \{-1, +1\}$. We use a very large weight on $y$ to emphasize its role in
determining the distance between data points, i.e., for a pair $(\bx_i, y_i)$ and
$(\bx_j, y_j)$, if $y_i \neq y_j$, they are considered to be infinitely far away from
each other; otherwise their distance is determined solely by the predictors. Our
robust LG problem is modeled as:
\begin{equation} \label{dro-lg}
\inf\limits_{\bbeta}\sup\limits_{\mbb{Q}\in \Omega}
\mbb{E}^{\mbb{Q}}\big[ \log(1+\exp(-y \bbeta'\bx))\big], 
\end{equation}
where $\mbb{Q}$ is the probability distribution of $(\bx, y)$, belonging to some set
$\Omega$ that includes all probability distributions whose order-one Wasserstein
distance (defined on the metric space $(\scrZ, s)$) to the empirical distribution
$\hat{\mathbb{P}}_N$ is no more than $\epsilon$. In the following theorem, we
reformulate (\ref{dro-lg}) as a penalized empirical loss minimization problem.
\begin{thm} \label{dro-lg-reform}
	Suppose we observe $N$ realizations of the data, denoted by $(\bx_i, y_i)$,
	$i=1,\ldots,N$. When the Wasserstein metric is induced by (\ref{metric-lg}),
	the DRO problem (\ref{dro-lg}) can be reformulated as:
	\begin{equation} \label{convex-lg}
	\inf\limits_{\bbeta} \mbb{E}^{\hat{\mbb{P}}_N}\big[ l_{\bbeta}(\bx, y)\big] + \epsilon \|\bbeta\|_* = \inf\limits_{\bbeta} \frac{1}{N} \sum_{i=1}^N \log\bigl(1+\exp(-y_i \bbeta'\bx_i)\bigr) + \epsilon \|\bbeta\|_*.
	\end{equation}
\end{thm}

We note that \cite{shafieezadeh2015distributionally,shafieezadeh2017regularization,gao2017wasserstein} arrive at a similar formulation to (\ref{convex-lg}) by other
means of derivation. Different from these existing works, we will consider
specifically the application of (\ref{convex-lg}) to grouped predictors where the
goal is to induce group level sparsity on the coefficients/classifier. As in Section
\ref{sec:gwgl_lr}, we assume that the predictor vector $\bx$ can be decomposed into
$L$ groups, i.e., $\bx = (\bx^1, \ldots, \bx^L)$, each $\bx^l$ containing $p_l$
predictors of group $l$, and $\sum_{l=1}^L p_l = p$ (no overlap among groups). To
reflect the group sparse structure, we consider the $(2, \infty)$-norm of the
weighted predictor vector $\bx_{\bw} \triangleq (\frac{1}{\sqrt{p_1}}\bx^1, \ldots,
\frac{1}{\sqrt{p_L}}\bx^L)$, where the weight vector is $\bw=(\frac{1}{\sqrt{p_1}},
\ldots, \frac{1}{\sqrt{p_L}})$. According to Theorem \ref{dualnorm}, the dual norm
of the weighted $(2, \infty)$-norm with weight $\bw=(\frac{1}{\sqrt{p_1}}, \ldots,
\frac{1}{\sqrt{p_L}})$ evaluated at $\bbeta$ is:
\begin{equation*}
\|\bbeta_{\bw^{-1}}\|_{2, 1} = \sum_{l=1}^L \sqrt{p_l}\|\bbeta^l\|_2,
\end{equation*} 
where $\bw^{-1} = (\sqrt{p_1}, \ldots, \sqrt{p_L})$, and $\bbeta^l$ denotes the vector of coefficients corresponding to group $l$. Therefore, the GWGL formulation for LG (GWGL-LG) takes the form: 
\begin{equation} \label{gwgl-lg} 
\inf\limits_{\bbeta} 
\frac{1}{N} \sum_{i=1}^N \log\bigl(1+\exp(-y_i \bbeta'\bx_i)\bigr) + \epsilon  
\sum_{l=1}^L \sqrt{p_l}\|\bbeta^l\|_2.
\end{equation}

The above derivation techniques also apply to other loss functions whose growth rate
is finite, e.g., the hinge loss used by the {\em Support Vector Machine (SVM)}, and
therefore, the GWGL SVM model can be developed in a similar fashion. It is also worth
noting that the regularizer in our tractable reformulation~(\ref{convex-lg}) is
related to the growth rate of the loss function, with the magnitude of the penalty
being the radius of the Wasserstein ball \cite{chen2017outlier,gao2017wasserstein}. This enables new perspectives of the regularization term and
provides guidance on the selection/tuning of the regularization coefficient.

\subsection{GLASSO with Overlapping Groups} \label{overlap}
In this subsection we will explore the GLASSO formulation with overlapping groups,
and show that our Wasserstein DRO framework recovers a latent GLASSO approach that was first
proposed in \cite{obozinski2011group}.

When the groups overlap with each other, the penalty term $\sum_{l=1}^L
\sqrt{p_l}\|\bbeta^l\|_2$ leads to a
solution whose support is almost surely the complement of a union of groups
\cite{jenatton2011structured}. In other words, setting one group to zero shrinks its
covariates to zero even if they belong to other groups, in which case these other
groups will not be entirely selected. \cite{obozinski2011group} proposed a latent
GLASSO approach where they introduce a set of latent variables that induce a solution
vector whose support is a union of groups, so that the estimator would select entire
groups of covariates. Specifically, define the latent variables $\bv^l \in \mbb{R}^p,
l=1, \ldots, L$ such that $\text{supp}(\bv^l) \subset \scrG^l, l=1, \ldots, L$, where
$\text{supp}(\bv^l) \subset \{1, \ldots, p\}$ denotes the support of $\bv^l$, i.e.,
the set of predictors $i \in \{1, \ldots, p\}$ such that $v_i^l \neq 0$, and
$\scrG^l$ denotes the set of predictors that are in group $l$. Our assumption is that
$\exists \ l_1, l_2$ such that $\scrG^{l_1} \cap \scrG^{l_2} \neq \emptyset$. The
latent GLASSO formulation has the form:
\begin{equation} \label{o1}
\begin{aligned}
& \inf\limits_{\bbeta, \bv^1, \ldots, \bv^L} \quad
\frac{1}{N}\sum\limits_{i=1}^N l_{\bbeta}(\bx_i, y_i)+ \epsilon  
\sum_{l=1}^L d_l \|\bv^l\|_2, \\
& \quad \ \text{s.t.} \ \qquad \bbeta = \sum_{l=1}^L \bv^l,
\end{aligned}
\end{equation}
where $l_{\bbeta}(\bx_i, y_i)$ denotes the loss at sample $(\bx_i, y_i)$, and $d_l$ is a user-specified penalty strength of group $l$.
Let $\hat{\bv}^l$, $l=1,\ldots,L$, denote an optimal solution of (\ref{o1}).  
By
using the latent vectors, Formulation (\ref{o1}) has the flexibility of
implicitly adjusting the support of the latent vectors such that for any $i \in
\text{supp}(\hat{\bv}^l)$ where $\hat{\bv}^l = \mathbf{0}$, it does not belong to the
support of any non-shrunk latent vectors. As a result, the covariates that belong to both
shrunk and non-shrunk groups would not be mistakenly driven to zero. 
Formulation (\ref{o1}) favors solutions which shrink some $\bv^l$ to zero,
while the non-shrunk components satisfy $\text{supp}(\bv^l) = \scrG^l$, therefore
leading to estimators whose support is the union of a set of groups.

To show that (\ref{o1}) can be obtained from the Wasserstein DRO framework, we
consider the following weighted $(2, \infty)$-norm on the predictor space:
\begin{equation} \label{overlapmetric}
s(\bx) = \max_l d_l^{-1} \|\bx^l\|_2.
\end{equation}
For simplicity we treat the response $y$ as a deterministic quantity so that the
Wasserstein metric is defined only on the predictor space. The scenario with
stochastic responses can be accommodated by introducing some constant
$M$. \cite{obozinski2011group} showed that the dual norm of (\ref{overlapmetric}) is
$\Omega(\bbeta) \triangleq \sum_{l=1}^L d_l \|\bv^l\|_2,$
with $\bbeta = \sum_{l=1}^L \bv^l$,
and $\bbeta \mapsto \Omega(\bbeta)$ is a valid norm.
By reformulating (\ref{o1}) as:
\begin{equation} \label{o2}
\inf\limits_{\bbeta} \quad
\frac{1}{N}\sum\limits_{i=1}^N l_{\bbeta}(\bx_i, y_i) + \epsilon  
\Omega(\bbeta),
\end{equation}
with 
\begin{equation*}
\Omega(\bbeta) = \min_{\substack{\bv^1, \ldots, \bv^L,\\ 
		\sum_{l=1}^L \bv^l= \bbeta}} \sum_{l=1}^L d_l \|\bv^l\|_2,
\end{equation*}
we have shown that (\ref{o1}) can be derived as a consequence of the Wasserstein DRO
formulation with the Wasserstein metric induced by (\ref{overlapmetric}). In fact,
(\ref{o2}) is equivalent to a regular
GLASSO in a covariate space of higher dimension obtained by duplication of the
covariates belonging to several groups. For simplicity our subsequent analysis
assumes non-overlapping groups.

\section{Grouping Effect of the Estimators} \label{s3} 
In this section we establish a {\em grouping effect} for the solutions to GWGL-LR and GWGL-LG, which measures the similarity of the
estimated coefficients in the same group. Ideally, for highly correlated predictors in
the same group, it is desired that their coefficients are close so that they can be
jointly selected/dropped (group sparsity).
The discussion on the prediction and estimation quality of the solutions is deferred to Appendix A.

To investigate the grouping effect of the
estimators, we examine the difference between coefficient estimates as a function of the sample correlation between their
corresponding predictors in the following theorem. 
\begin{thm} \label{grouping2} Suppose the predictors are
	standardized (columns of $\bX$ have zero mean and unit variance). Let $\hat{\bbeta}
	\in \mbb{R}^p$ be the optimal solution to (\ref{gwgl-lr}) (or (\ref{gwgl-lg})).  If $\bx_{,i}$
	is in group $l_1$ and $\bx_{,j}$ is in group $l_2$, and
	$\|\hat{\bbeta}^{l_1}\|_2 \neq 0$, $\|\hat{\bbeta}^{l_2}\|_2 \neq 0$,
	define
	\begin{equation*}
	D(i, j) =
	\Biggl|\frac{\sqrt{p_{l_1}}\hat{\beta}_i}{\|\hat{\bbeta}^{l_1}\|_2}
	-\frac{\sqrt{p_{l_2}}\hat{\beta}_j}{\|\hat{\bbeta}^{l_2}\|_2}\Biggr|. 
	\end{equation*} 
	Then,
	$$D(i, j) \le \frac{\sqrt{2(1-\rho)}}{\sqrt{N}\epsilon},$$
	where $\rho = \bx_{,i}'\bx_{,j}$ is the sample correlation, and
	$p_{l_1}, p_{l_2}$ are the number of predictors in groups $l_1$ and
	$l_2$, respectively.
\end{thm}

Theorem \ref{grouping2} establishes a unified result for the grouping effect of the GWGL-LR and GWGL-LG solutions. When $\bx_{,i}$ and $\bx_{,j}$ are both in group $l$ and $\|\hat{\bbeta}^{l}\|_2
\neq 0$, it follows
\begin{equation} \label{group1}
|\hat{\beta}_i - \hat{\beta}_j| \le
\frac{\sqrt{2(1-\rho)}\|\hat{\bbeta}^{l}\|_2}{\epsilon \sqrt{N p_l}}.
\end{equation}
From (\ref{group1}) we see that as the within group correlation increases, the
difference between $\hat{\beta}_i$ and $\hat{\beta}_j$ becomes smaller. In the
extreme case where $\bx_{,i}$ and $\bx_{,j}$ are perfectly correlated, $\hat{\beta}_i = \hat{\beta}_j$. This grouping effect enables recovery of
sparsity on a group level when the correlation between predictors in the same group
is high, and implies the use of predictors' correlation as a grouping criterion. One
of the popular clustering algorithms, called {\em spectral clustering}
\cite{shi2000normalized,ng2002spectral}, performs grouping based on the
eigenvalues/eigenvectors of the Laplacian matrix of the similarity graph that is
constructed using the {\em similarity matrix} of data, and divides the data points
(predictors) into several groups such that points in the same group are similar and
points in different groups are dissimilar to each other. The similarity matrix
measures the pairwise similarities between data points, which in our case could be
the pairwise correlations between predictors.


\section{Numerical Results} \label{s4} 

In this section we compare our GWGL formulations with other commonly used predictive
models. In the linear regression setting, we compare GWGL-LR with models that either
(i) use a different loss function, e.g., the traditional GLASSO with an $\ell_2$-loss
\cite{yuan2006model}, and the Group Square-Root LASSO (GSRL) \cite{bunea2014group}
that minimizes the square root of the $\ell_2$-loss; or (ii) do not make use of the
grouping structure of the predictors, e.g., the Elastic Net (EN)
\cite{zou2005regularization}, and the LASSO \cite{tibshirani1996regression}.  For
classification problems, we consider alternatives that minimize the empirical logloss
plus penalty terms that do not utilize the grouping structure of the predictors,
e.g., the $\ell_1$-regularizer (LG-LASSO), $\ell_2$-regularizer (LG-Ridge), and their
combination (LG-EN).  

\subsection{GWGL-LR on Synthetic Datasets} \label{gwgl-lr-exp}
In this subsection we will compare GWGL-LR with the aforementioned models on several
synthetic datasets. The data is generated as follows:
\begin{enumerate}
	\item Set $\beta^*_i$ to $0.5$ if predictor $i$ belongs to an even group, and
	$0$ otherwise.
	\item Generate $\bx \in \mbb{R}^{p}$ from the Gaussian distribution $\scrN_p
	(0, \bSigma)$, where $\bSigma=(\sigma_{i,j})_{i,j=1}^p$ has diagonal
	elements equal to $1$, and off-diagonal elements $\sigma_{i,j}$ equal to
	$\rho_w$ if predictors $i$ and $j$ are in the same group, and $0$
	otherwise. Here $\rho_w$ is called the {\em within group correlation}. 
	\item With probability $1-q$, generate $y$ from $\scrN(\bx'\bbeta^*, \sigma^2)$, and with probability $q$, generate $y$ from $\scrN(\bx'\bbeta^*+ 5\sigma, \sigma^2)$,
	where $\sigma^2$ is the intrinsic variance of $y$, and $q$ is the probability of abnormal
	samples (outliers).
\end{enumerate}

We generate 10 datasets consisting of $N = 100$ training samples and $M_t = 60$ test
samples with 4 groups of predictors, where $p_1
= 1, p_2 = 3, p_3 = 5, p_4 = 7$, and $p=\sum_{i=1}^4 p_i=16$. We are interested in
studying the impact of (i) {\em Signal to Noise Ratio (SNR)}, defined as: $\text{SNR}
= (\bbeta^*)'\bSigma \bbeta^*/\sigma^2$, and (ii) the {\em within group correlation}
$\rho_w$.  The performance metrics we use include (i) {\em Median Absolute Deviation
	(MAD)} on the test set, which is defined to be the median value of $|y_i -
\bx_i'\hat{\bbeta}|, \ i=1, \ldots, M_t$, with $\hat{\bbeta}$ being the estimate of
$\bbeta^*$ obtained from the training set, and $(\bx_i, y_i), \ i=1, \ldots, M_t,$
being the observations from the test set; (ii) {\em Relative Risk (RR)}, {\em
	Relative Test Error (RTE)}, and {\em Proportion of Variance Explained (PVE)} of
$\hat{\bbeta}$ (see definitions in Appendix B).

Before solving for the regression coefficients, the grouping of predictors needs to
be determined. Unlike most of the existing works where the grouping structure is
assumed to be known a priori, we propose to use a data-driven clustering
algorithm to group the predictors based on their sample correlations. Specifically, we consider the {\em spectral clustering}
\cite{shi2000normalized,ng2002spectral} algorithm with the Gaussian
similarity function
$\text{Gs}(\bx_{,i}, \bx_{,j}) \triangleq \exp\big( -\|\bx_{,i} - \bx_{,j}\|_2^2/(2\sigma_s^2)\big)$ that captures the sample pairwise correlations between
predictors, where $\sigma_s$ is some scale parameter whose selection is discussed in Appendix B.

We plot two sets of graphs: (i) the performance metrics v.s. SNR, where SNR is equally spaced between 0.5 and 2 on a log scale, and $\rho_w$ is set to
$0.8$ times a random noise uniformly distributed on the interval $[0.2, 0.4]$; and
(ii) the performance metrics v.s. $\rho_w$, where
$\rho_w$ takes values in $(0.1, 0.2, \ldots, 0.9)$, and SNR is fixed to $1$.  In the
graphs for RR, RTE and PVE, we also plot the ideal scores, which are the values
achieved by $\hat{\bbeta} = \bbeta^*$, and the null scores, which are the values
achieved by $\hat{\bbeta} = 0$.
We only show results for $q =
30\%$. The figures for $q =
20\%$ can be found in Appendix B.
\begin{figure}[ht] 
	\begin{subfigure}{.49\textwidth}
		\centering
		\includegraphics[width=0.9\textwidth, height = 2.4in]{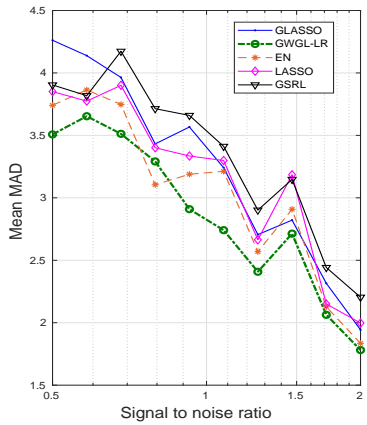}
		\caption{\small{Median Absolute Deviation.}}
	\end{subfigure}
	\begin{subfigure}{0.49\textwidth}
		\centering
		\includegraphics[width=0.9\textwidth, height = 2.4in]{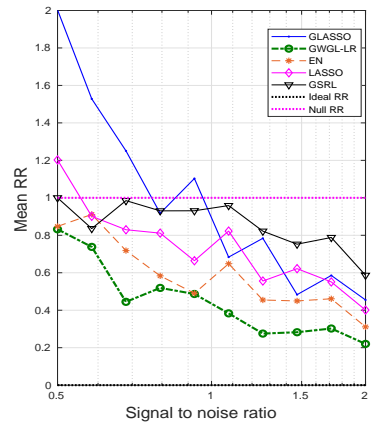}
		\caption{\small{Relative risk.}}
	\end{subfigure}
	
	\begin{subfigure}{0.49\textwidth}
		\centering
		\includegraphics[width=0.9\textwidth, height = 2.4in]{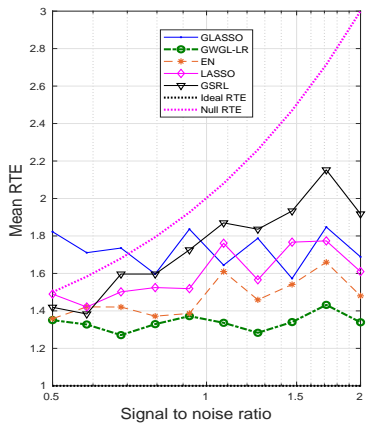}
		\caption{\small{Relative test error.}}
	\end{subfigure}%
	\begin{subfigure}{0.49\textwidth}
		\centering
		\includegraphics[width=0.9\textwidth, height = 2.4in]{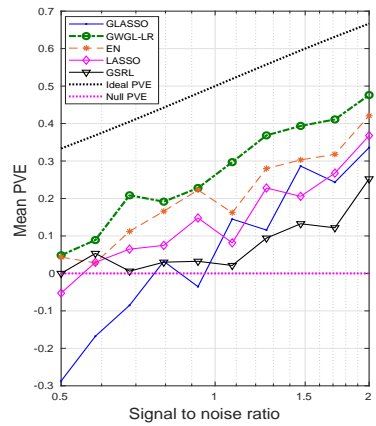}
		\caption{\small{Proportion of variance explained.}}
	\end{subfigure}
	\caption{The impact of SNR on the performance metrics, $q =
		30\%$.} \label{snr-30}
\end{figure}

\begin{figure}[h] 
	\begin{subfigure}{.49\textwidth}
		\centering
		\includegraphics[width=0.9\textwidth]{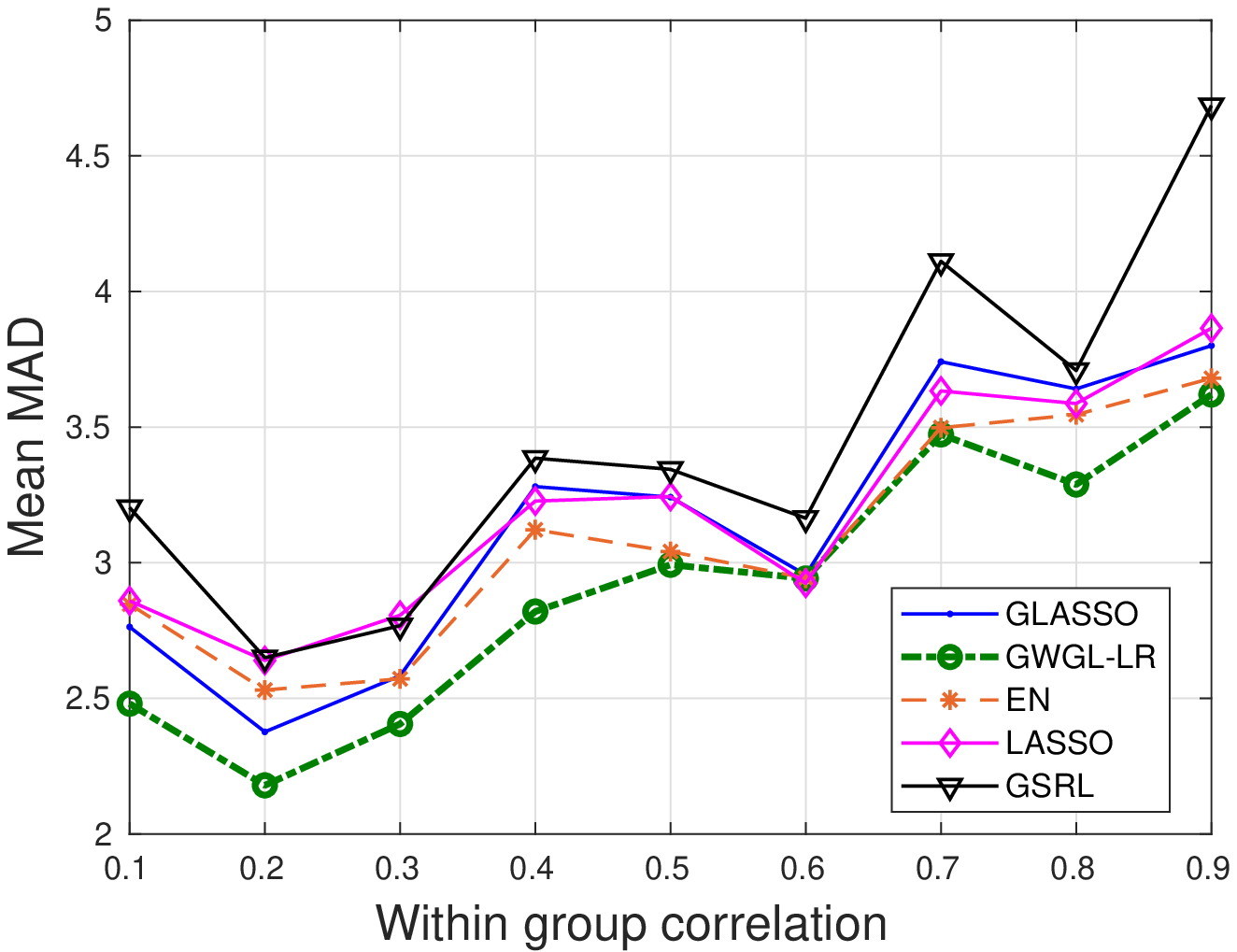}
		\caption{\small{Median Absolute Deviation.}}
	\end{subfigure}
	\begin{subfigure}{0.49\textwidth}
		\centering
		\includegraphics[width=0.9\textwidth]{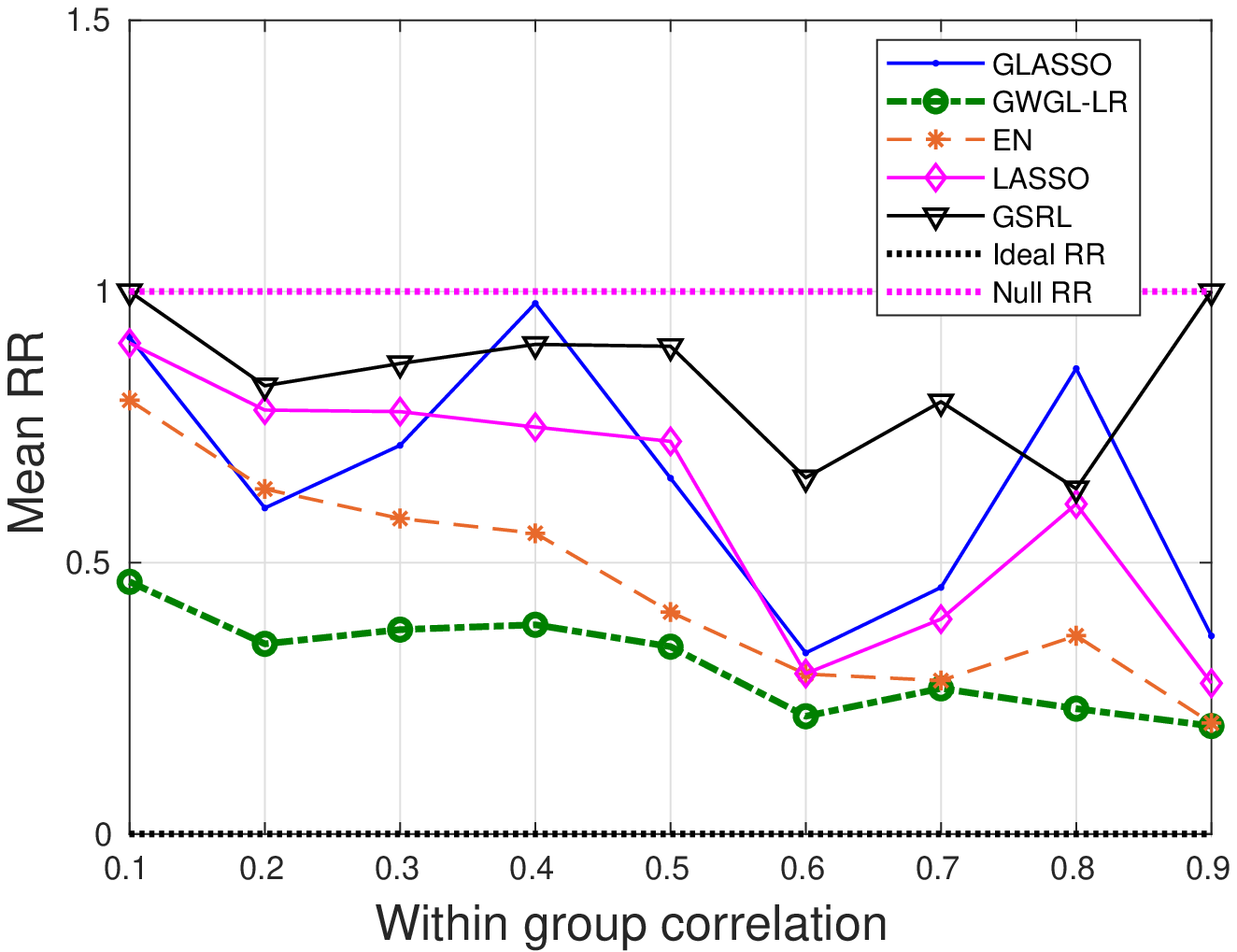}
		\caption{\small{Relative risk.}}
	\end{subfigure}
	
	\begin{subfigure}{0.49\textwidth}
		\centering
		\includegraphics[width=0.9\textwidth]{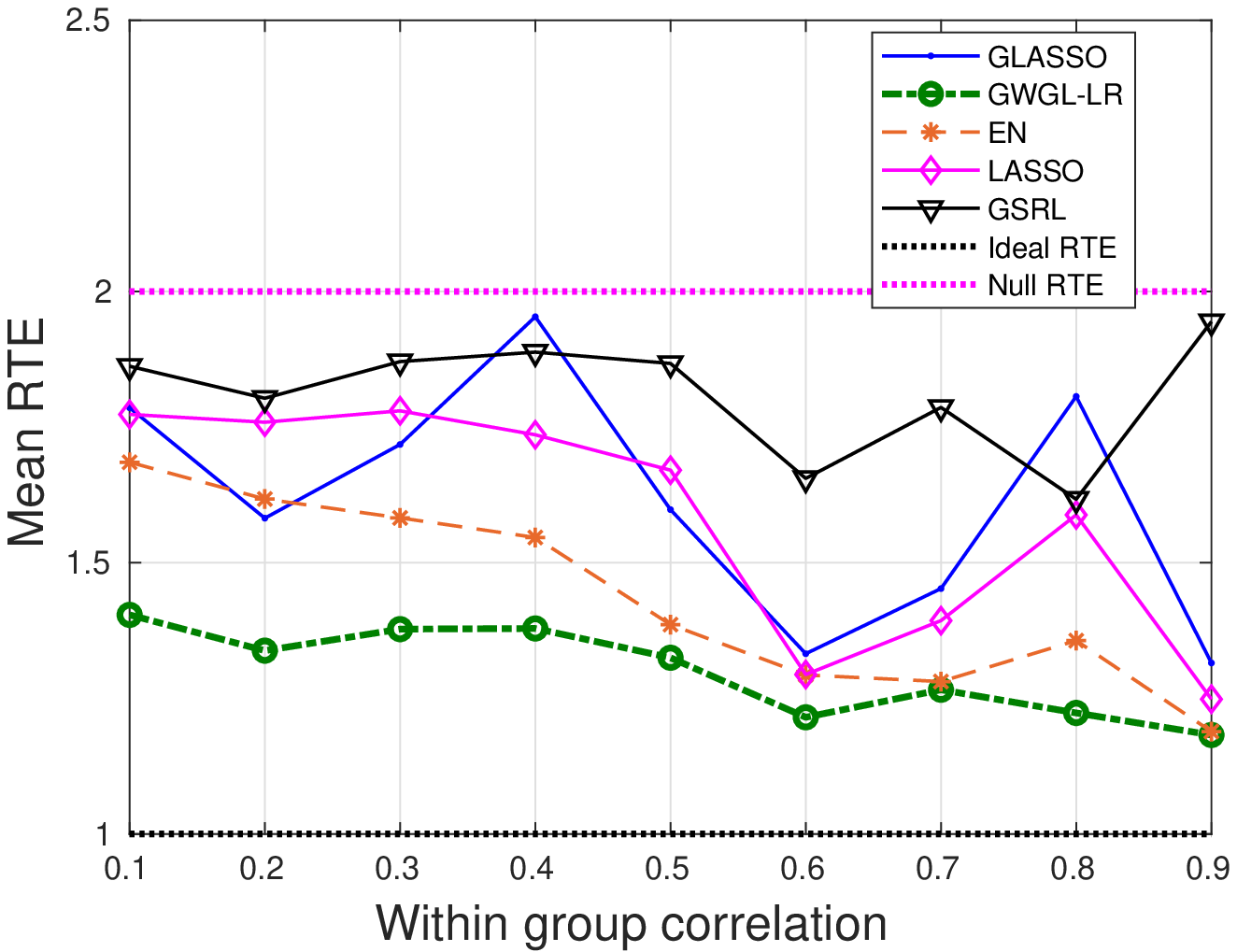}
		\caption{\small{Relative test error.}}
	\end{subfigure}%
	\begin{subfigure}{0.49\textwidth}
		\centering
		\includegraphics[width=0.9\textwidth]{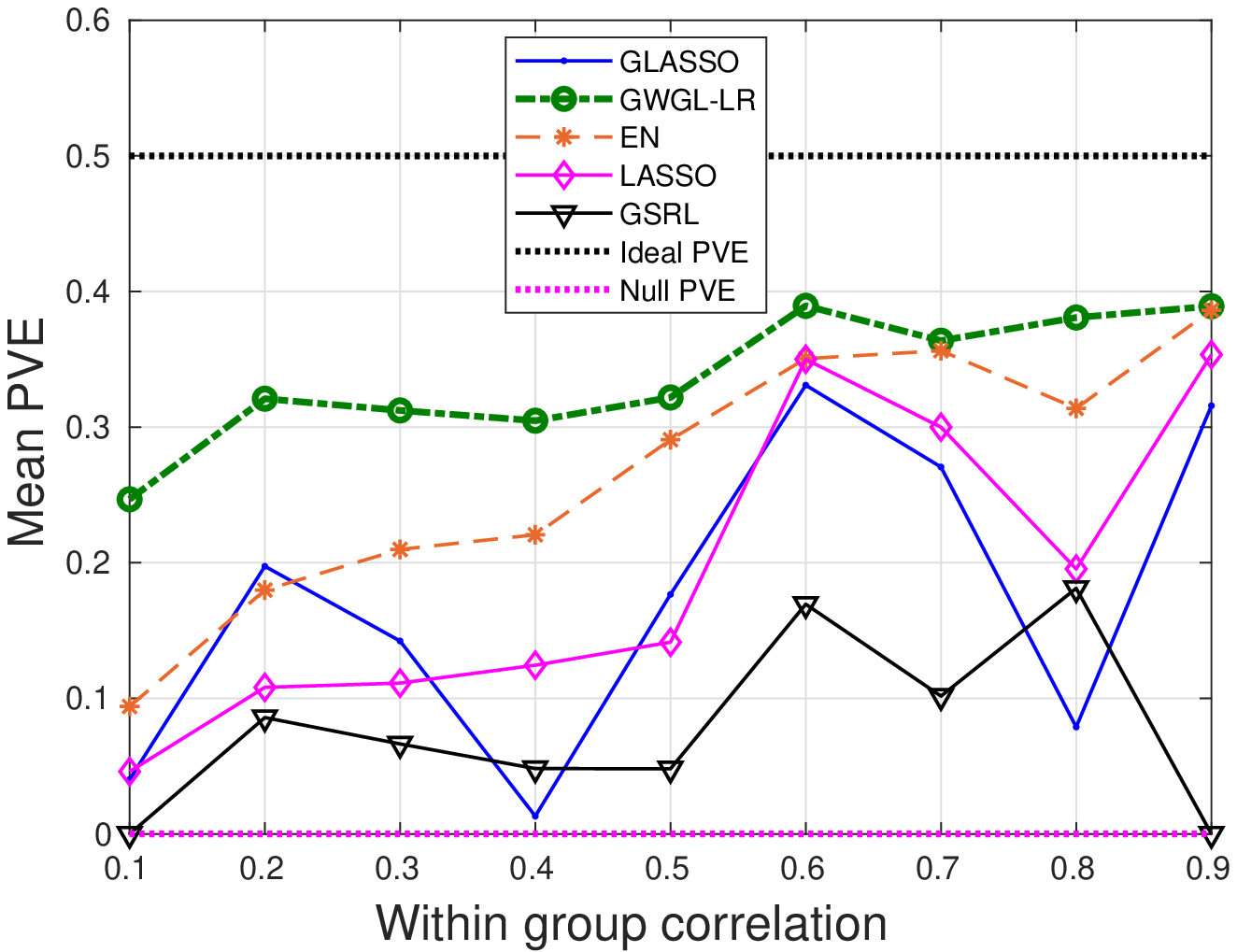}
		\caption{\small{Proportion of variance explained.}}
	\end{subfigure}
	\caption{{\small The impact of within group correlation on the performance metrics, $q =
		30\%$.}} \label{cor-30}
\end{figure}

To better highlight the benefits of GWGL-LR, we define the {\em Maximum Percentage Improvement (MPI)} to be the maximum percentage difference of the performance metrics between GWGL-LR and
the best among all others. The MPI values for all metrics are shown in Tables~\ref{table1} and \ref{table2} in Appendix B.

We summarize below our main findings from the results we have presented: (i) for all approaches, MAD and RR decrease as the data becomes less noisy. PVE increases when the noise is reduced; (ii) the GWGL-LR formulation has better prediction and estimation
performance than all other approaches under consideration. When the within
group correlation is varied, GWGL-LR shows a more stable performance; and (iii) the relative improvement of GWGL-LR over GLASSO 
is more significant for highly noisy data, which can be attributed to the
$\ell_1$-loss function it uses. Moreover, GWGL-LR generates more stable
estimators than GLASSO.

\subsection{Surgery Dataset} \label{surgery}

In this section we test our GWGL formulations on a real dataset obtained from the National Surgical Quality Improvement
Program (NSQIP) containing medical
records of patients who underwent a general surgical procedure. The dataset
includes (i) baseline demographics; (ii) pre-existing comorbidity information; (iii)
preoperative variables; (iv) index admission-related diagnosis and procedure
information; (v) postoperative events and complications, and (vi) additional
socioeconomic variables.

In our study, patients who underwent a general surgery procedure over 2011--2014 and
were tracked by the NSQIP were identified. We will focus on two supervised learning
models: (i) a linear regression model whose objective is to predict the
post-operative hospital length of stay, and
(ii) an LG model whose objective is to predict the re-hospitalization of patients
within 30 days after discharge. Both
models are extremely useful as they allow hospital staff to predict post-operative
bed occupancy and prevent costly 30-day readmissions.

The post-processed datasets include a total
of $2,275,452$ records, with $131$
numerical predictors for the regression model and $132$ for the classification model. The spectral clustering algorithm is used to group the predictors, with the number of groups specified as $67$ based on a
preliminary analysis.

For predicting the hospital length of stay, we report the mean (std.) of the
out-of-sample MAD across $5$ repetitions in Table~\ref{table7}. We see that our GWGL-LR formulation achieves the lowest mean
MAD with a small variation. Compared to the best among others, we improve the mean MAD by $7.30\%$. For longer hospital length of
stay, this could imply 1 or 2 days improvement in prediction accuracy, which is both
clinically and economically significant.

\begin{table}[ht]
	\caption{The mean and standard deviation of MAD on
		the surgery data.} \label{table7}
	\begin{center}
		{\footnotesize 	 \begin{tabular}{c|>{\centering\arraybackslash}p{1.5cm}|>{\centering\arraybackslash}p{1.5cm}|>{\centering\arraybackslash}p{1.5cm}|>{\centering\arraybackslash}p{1.5cm}|>{\centering\arraybackslash}p{1.5cm}} 
				\hline
				& GLASSO         & GWGL-LR            & EN          & LASSO     & GSRL  \\ \hline
				Mean (Std.) & 0.17 (0.0007)          & 0.16 (0.001)         & 0.17 (0.0009)      & 0.17 (0.0009) &  0.17 (0.0009)   \\ \hline
		\end{tabular}}
	\end{center}
\end{table}
	
For predicting the re-hospitalization of patients, we notice that the dataset
is highly unbalanced, with only $6\%$ of patients being re-hospitalized. To obtain a
balanced training set, we randomly draw $20\%$ patients from the positive class
(re-hospitalized patients), and sample the same number of patients from the negative
class, resulting in a training set of size $53,616$. All the remaining patients are
assigned to the test dataset. All formulations achieve an average {\em out-of-sample
ACC} (the prediction accuracy on the test dataset)
around $0.62$, an average {\em out-of-sample AUC} (Area Under the ROC Curve) of
$0.83$, and an average {\em logloss} on the test set ranging from $0.84$ to
$0.87$. We define a new performance metric, called the {\em Within Group Difference
(WGD)}, to measure the ability of the solution to induce group level sparsity.
\begin{equation*}
\text{WGD}(\hat{\bbeta}) \triangleq \frac{1}{|\{l: p_l \ge 2\}|} \sum_{l: p_l \ge 2} \frac{1}{\binom{p_l}{2}} \sum_{x_i, x_j \in \bx^l} \biggl|\frac{\hat{\beta}_i - \hat{\beta}_j}{\bx_{,i}'\bx_{,j}}\biggr|,
\end{equation*}
where $|\{l: p_l \ge 2\}|$ denotes the cardinality of the set $\{l: p_l \ge 2\}$, and
$\bx_{,i}'\bx_{,j}$ measures the sample correlation between predictors $x_i$ and
$x_j$. Theorem \ref{grouping-lg} implies that the higher the correlation,
the smaller the difference between the coefficients, and thus, a smaller WGD value
would suggest a stronger ability of grouped variable selection. Table
\ref{wgd-surgery} suggests that GWGL-LG encourages group level
sparsity. From Table \ref{dropgroup} we conclude that though LG-EN and
LG-LASSO obtain the most parsimonious model at an individual level, GWGL-LG has a stronger ability to induce
group level sparsity.

\begin{table}[ht]
\caption{The WGD of the estimators on the surgery data.} \label{wgd-surgery} 
\begin{center}
{\small \begin{tabular}{ c|>{\centering\arraybackslash}p{1.5cm}|>{\centering\arraybackslash}p{1.5cm}|>{\centering\arraybackslash}p{1.5cm}|>{\centering\arraybackslash}p{1.5cm}|>{\centering\arraybackslash}p{1.5cm} } 
		\hline
		& LG            & LG-LASSO      & LG-Ridge      & LG-EN          & GWGL-LG \\ \hline 
		Mean (Std.) & 23.93 (1.28)  & 16.28 (0.72)  & 23.38 (1.15)  &  16.26 (0.74)  & 5.04 (0.45)  \\ 
		\hline
	\end{tabular}}
\end{center}
\end{table}

\begin{table}[ht]
\caption{The number of dropped groups/features on the surgery data.} \label{dropgroup} 
\begin{center}
	{\small \begin{tabular}{ c|>{\centering\arraybackslash}p{0.6cm}|>{\centering\arraybackslash}p{1.5cm}|>{\centering\arraybackslash}p{1.3cm}|>{\centering\arraybackslash}p{1.2cm}|>{\centering\arraybackslash}p{1.5cm} } 
			\hline
			& LG & LG-LASSO   & LG-Ridge   & LG-EN & GWGL-LG \\ \hline 
			No. of dropped groups  & 1  & 6          & 2          &  10   & 16 \\ 
			No. of dropped features& 2  & 24         & 2          &  25   & 19\\ 
			\hline
		\end{tabular}}
	\end{center}
\end{table}

\section{Conclusions} \label{s5} 

We proposed a DRO formulation under the Wasserstein metric that recovers the GLASSO
penalty for LAD and LG, through which we have established a connection between
group-sparse regularization and robustness. We provided insights on the grouping effect of our estimators,
which suggests the use of spectral clustering with the Gaussian similarity function
to perform grouping on the predictors.
We reported results from several experiments, showing that our
formulations achieve more accurate and stable estimates, and 
have a stronger ability of inducing group level sparsity.

\begin{acknowledgements}
We thank George Kasotakis, MD, MPH, for providing access to the surgery dataset. We
also thank Taiyao Wang for help in processing this dataset.  Research was partially
supported by the NSF under grants IIS-1914792, DMS-1664644, and CNS-1645681, by the
NIH under grant 1R01GM135930, and by the ONR under grant N00014-19-1-2571.
\end{acknowledgements}

\appendix  
\section*{Appendix A: Omitted Theoretical Results and Proofs} 

This section contains the theoretical statements and proofs that are omitted in Sections \ref{s2} and \ref{s3}. 

\subsection*{Proof of Theorem \ref{mixture}}
\begin{proof} 
	From the definition of the Wasserstein distance, $W_1 (\mathbb{P}_{\text{out}}, \mathbb{P}_{\text{mix}})$ is the optimal value of the following optimization problem:
	\begin{equation} \label{wassQ1}
	\begin{aligned}
	\min\limits_{\Pi \in \scrP(\scrZ \times \scrZ)} & \quad \int_{\scrZ \times \scrZ} s(\bz_1, \bz_2) \ \Pi \bigl(d\bz_1, d\bz_2\bigr) \\
	\text{s.t.} & \quad \int_{\scrZ}\Pi \bigl(\bz_1, \bz_2\bigr) d\bz_2 = \mathbb{P}_{\text{out}}(\bz_1), \forall \bz_1 \in \scrZ, \\
	& \quad \int_{\scrZ}\Pi \bigl(\bz_1, \bz_2\bigr) d\bz_1 = q \mathbb{P}_{\text{out}}(\bz_2) + (1-q)\mathbb{P}(\bz_2), \forall \bz_2 \in \scrZ.
	\end{aligned}
	\end{equation}
	Similarly, $W_1 (\mathbb{P}, \mathbb{P}_{\text{mix}})$ is the optimal value of the following optimization problem:
	\begin{equation} \label{wassQ2}
	\begin{aligned}
	\min\limits_{\Pi \in \scrP(\scrZ \times \scrZ)} & \quad \int_{\scrZ \times \scrZ} s(\bz_1, \bz_2) \ \Pi \bigl(d\bz_1, d\bz_2\bigr) \\
	\text{s.t.} & \quad \int_{\scrZ}\Pi \bigl(\bz_1, \bz_2\bigr) d\bz_2 = \mathbb{P}(\bz_1), \forall \bz_1 \in \scrZ, \\
	& \quad \int_{\scrZ}\Pi \bigl(\bz_1, \bz_2\bigr) d\bz_1 = q \mathbb{P}_{\text{out}}(\bz_2) + (1-q)\mathbb{P}(\bz_2), \forall \bz_2 \in \scrZ.
	\end{aligned}
	\end{equation}
	
	We propose a decomposition strategy. For Problem (\ref{wassQ1}), decompose the joint distribution $\Pi$ as $\Pi = (1-q)S + q T$, where $S$ and $T$ are two joint distributions of $\bz_1$ and $\bz_2$. The first set of constraints in Problem (\ref{wassQ1}) can be equivalently expressed as:
	\begin{equation*} 
	(1-q)\int_{\scrZ}S\bigl(\bz_1, \bz_2\bigr)d\bz_2 + q \int_{\scrZ}T\bigl(\bz_1, \bz_2\bigr)  d\bz_2 = (1-q)\mathbb{P}_{\text{out}}(\bz_1) + q \mathbb{P}_{\text{out}}(\bz_1), \forall \bz_1 \in \scrZ,
	\end{equation*}
	and thus,
	$$\int_{\scrZ}S\bigl(\bz_1, \bz_2\bigr)d\bz_2 = \mathbb{P}_{\text{out}}(\bz_1), \quad \int_{\scrZ}T\bigl(\bz_1, \bz_2\bigr)d\bz_2 = \mathbb{P}_{\text{out}}(\bz_1), \forall \bz_1 \in \scrZ.$$	
	The second set of constraints can be expressed as:
	\begin{equation*}
	(1-q)\int_{\scrZ}S\bigl(\bz_1, \bz_2\bigr)d\bz_1 + q \int_{\scrZ}T\bigl(\bz_1, \bz_2\bigr)  d\bz_1 = q \mathbb{P}_{\text{out}}(\bz_2) + (1-q)\mathbb{P}(\bz_2), \forall \bz_2 \in \scrZ,
	\end{equation*}
	which implies that
	$$\int_{\scrZ}S\bigl(\bz_1, \bz_2\bigr)d\bz_1 = \mathbb{P}(\bz_2), \quad \int_{\scrZ}T\bigl(\bz_1, \bz_2\bigr) d\bz_1 = \mathbb{P}_{\text{out}}(\bz_2), \forall \bz_2 \in \scrZ.$$
	The objective function can be decomposed as:
	\begin{equation*}
	\begin{aligned}
	\int_{\scrZ \times \scrZ} s(\bz_1, \bz_2) \ \Pi \bigl(d\bz_1, d\bz_2\bigr)  = & \ (1-q)\int_{\scrZ \times \scrZ} s(\bz_1, \bz_2) \ S\bigl(d\bz_1, d\bz_2\bigr) \\
	& + q \int_{\scrZ \times \scrZ} s(\bz_1, \bz_2)T \bigl(d\bz_1, d\bz_2\bigr).
	\end{aligned}
	\end{equation*}
	Therefore, Problem (\ref{wassQ1}) can be decomposed into the following two subproblems.
	\[ \text{Subproblem 1:} \qquad \begin{array}{rl}
	\min\limits_{S \in \scrP(\scrZ \times \scrZ)} &  \int_{\scrZ \times \scrZ} s(\bz_1, \bz_2) \ S\bigl(d\bz_1, d\bz_2\bigr) \\
	\text{s.t.} &  \int_{\scrZ}S\bigl(\bz_1, \bz_2\bigr)d\bz_2 = \mathbb{P}_{\text{out}}(\bz_1), \forall \bz_1 \in \scrZ,\\
	& \int_{\scrZ}S\bigl(\bz_1, \bz_2\bigr)d\bz_1 = \mathbb{P}(\bz_2), \forall \bz_2 \in \scrZ.
	\end{array}
	\]
	\[ \text{Subproblem 2:} \qquad \begin{array}{rl}
	\min\limits_{T \in \scrP(\scrZ \times \scrZ)} &  \int_{\scrZ \times \scrZ} s(\bz_1, \bz_2) \ T\bigl(d\bz_1, d\bz_2\bigr) \\
	\text{s.t.} &  \int_{\scrZ}T\bigl(\bz_1, \bz_2\bigr)d\bz_2 = \mathbb{P}_{\text{out}}(\bz_1), \forall \bz_1 \in \scrZ,\\
	& \int_{\scrZ}T\bigl(\bz_1, \bz_2\bigr)d\bz_1 = \mathbb{P}_{\text{out}}(\bz_2), \forall \bz_2 \in \scrZ.
	\end{array}
	\]		
	Assume that the optimal solutions to the two subproblems are $S^*$ and $T^*$, respectively, we know $\Pi_0 = (1-q)S^* + q T^*$ is a feasible solution to Problem (\ref{wassQ1}). Therefore,
	\begin{equation} \label{out-mix}
	\begin{aligned}
	W_1 (\mathbb{P}_{\text{out}}, \mathbb{P}_{\text{mix}}) & \le \int_{\scrZ \times \scrZ} s(\bz_1, \bz_2) \ \Pi_0 \bigl(d\bz_1, d\bz_2\bigr) \\
	& = (1-q)W_1 (\mathbb{P}_{\text{out}}, \mathbb{P}) + q W_1 (\mathbb{P}_{\text{out}}, \mathbb{P}_{\text{out}})\\
	& = (1-q)W_1 (\mathbb{P}_{\text{out}}, \mathbb{P}).
	\end{aligned}
	\end{equation}
	Similarly, 
	\begin{equation} \label{true-mix}
	W_1 (\mathbb{P}, \mathbb{P}_{\text{mix}}) \le q W_1 (\mathbb{P}_{\text{out}}, \mathbb{P}).
	\end{equation}
	(\ref{out-mix}) and (\ref{true-mix}) imply that
	\begin{equation*}
	W_1 (\mathbb{P}_{\text{out}}, \mathbb{P}_{\text{mix}}) + W_1 (\mathbb{P}, \mathbb{P}_{\text{mix}}) \le W_1 (\mathbb{P}_{\text{out}}, \mathbb{P}).
	\end{equation*}
	On the other hand, based on the subadditivity of the Wasserstein metric, we have,
	$$W_1 (\mathbb{P}_{\text{out}}, \mathbb{P}_{\text{mix}}) + W_1 (\mathbb{P}, \mathbb{P}_{\text{mix}}) \ge W_1 (\mathbb{P}_{\text{out}}, \mathbb{P}).$$
	We thus conclude that 
	\begin{equation} \label{equal}
	W_1 (\mathbb{P}_{\text{out}}, \mathbb{P}_{\text{mix}}) + W_1 (\mathbb{P}, \mathbb{P}_{\text{mix}}) = W_1 (\mathbb{P}_{\text{out}}, \mathbb{P}).
	\end{equation}
	To achieve the equality in (\ref{equal}), (\ref{out-mix}) and (\ref{true-mix}) must be equalities, i.e.,
	$$W_1 (\mathbb{P}_{\text{out}}, \mathbb{P}_{\text{mix}}) = (1-q)W_1 (\mathbb{P}_{\text{out}}, \mathbb{P}),$$
	and, 
	$$W_1 (\mathbb{P}, \mathbb{P}_{\text{mix}}) = q W_1 (\mathbb{P}_{\text{out}}, \mathbb{P}). $$
	Thus,
	\begin{equation*}
	\frac{W_1 (\mathbb{P}_{\text{out}}, \mathbb{P}_{\text{mix}})}{W_1 (\mathbb{P}, \mathbb{P}_{\text{mix}})} = \frac{(1-q)W_1 (\mathbb{P}_{\text{out}}, \mathbb{P})}{q W_1 (\mathbb{P}_{\text{out}}, \mathbb{P})} 
	=  \frac{1-q}{q}.
	\end{equation*} 
	\qed
\end{proof} 

\subsection*{Proof of Theorem \ref{dualnorm}}
\begin{proof} We will use H\"{o}lder's inequality, which we state for
	convenience.
	
	H\"{o}lder's inequality: Suppose we have two scalars $p, q \geq 1$ and $1/p
	+ 1/q =1$. For any two vectors $\ba = (a_1, \ldots, a_n)$ and $\bb =
	(b_1, \ldots, b_n)$, 
	\begin{equation*}
	\sum_{i=1}^n |a_i b_i| \le \Bigl(\sum_{i=1}^n |a_i|^p\Bigr)^{1/p}
	\Bigl(\sum_{i=1}^n |b_i|^q\Bigr)^{1/q}. 
	\end{equation*}
	
	The dual norm of $\|\cdot\|_{r, s}$ evaluated at some vector $\bbeta$ is the optimal value of problem (\ref{dualnorm2}):
	\begin{equation} \label{dualnorm2}
	\begin{aligned}
	\max\limits_{\bx} & \quad \bx' \bbeta \\
	\text{s.t.} & \quad \|\bx_{\bw}\|_{r, s} \le 1.
	\end{aligned}
	\end{equation}
	We assume
	that $\bbeta$ has the same group structure with $\bx$, i.e., $\bbeta =
	(\bbeta^1, \ldots, \bbeta^L)$. Using H\"{o}lder's inequality, we can
	write
	\begin{equation*}
	\bx'\bbeta = \sum_{l=1}^L (w_l\bx^l)'\Bigl(\frac{1}{w_l}\bbeta^l\Bigr)
	\le \sum_{l=1}^L \|w_l\bx^l\|_r \left\|\frac{1}{w_l}\bbeta^l\right\|_q. 
	\end{equation*} 
	Define two new vectors in $\mbb{R}^L$
	\begin{equation*}
	\bx_{new} = (\|w_1\bx^1\|_r, \ldots, \|w_L\bx^L\|_r), \quad
	\bbeta_{new} = \left(\left\|\frac{1}{w_1}\bbeta^1\right\|_q, \ldots,
	\left\|\frac{1}{w_L}\bbeta^L\right\|_q\right). 
	\end{equation*}
	Applying H\"{o}lder's inequality again to $\bx_{new}$ and
	$\bbeta_{new}$, we obtain:
	\begin{equation*}
	\begin{aligned}
	\bx'\bbeta & \le \bx_{new}'\bbeta_{new} \\
	& \le \|\bx_{new}\|_s
	\|\bbeta_{new}\|_t \\
	& = \Bigl(\sum_{l=1}^L
	\bigl(\|w_l\bx^l\|_r\bigr)^s\Bigr)^{1/s} \left(\sum_{l=1}^L
	\left(\left\|\frac{1}{w_l}\bbeta^l\right\|_q\right)^t\right)^{1/t}. 
	\end{aligned}
	\end{equation*} 
	Therefore,
	\begin{equation*}
	\bx'\bbeta \le \|\bx_{\bw}\|_{r, s} \|\bbeta_{\bw^{-1}}\|_{q, t} \le
	\|\bbeta_{\bw^{-1}}\|_{q, t}, 
	\end{equation*}
	due to the constraint $\|\bx_{\bw}\|_{r, s} \le 1$. The result then
	follows.
	\qed
\end{proof}

\subsection*{Proof of Theorem \ref{dro-lg-reform}}
\begin{proof}
	To derive a tractable reformulation of the DRO-LG problem (\ref{dro-lg}), we borrow the idea from \cite{chen2017outlier} and \cite{gao2016distributionally}, which states that for any $\mbb{Q} \in \Omega$,
	\begin{equation} \label{finitedif} 
		\begin{aligned}
		& \Bigl|\mbb{E}^{\mbb{Q}}\big[ l_{\bbeta}(\bx, y)\big] -  \mbb{E}^{\hat{\mbb{P}}_N}\big[ l_{\bbeta}(\bx, y)\big]\Bigr|  \\
		= & \biggl|\int_{\scrZ} l_{\bbeta}(\bx_1, y_1) \mbb{Q}(d(\bx_1, y_1)) - \int_{\scrZ} l_{\bbeta}(\bx_2, y_2) \hat{\mbb{P}}_N(d(\bx_2, y_2)) \biggr| \\
		= & \biggl|\int_{\scrZ} l_{\bbeta}(\bx_1, y_1) \int_{\scrZ}\Pi_0(d(\bx_1, y_1), d(\bx_2, y_2)) - \int_{\scrZ} l_{\bbeta}(\bx_2, y_2) \int_{\scrZ}\Pi_0(d(\bx_1, y_1), d(\bx_2, y_2)) \biggr| \\
		\le & \int_{\scrZ \times \scrZ} \bigl|l_{\bbeta}(\bx_1, y_1)-l_{\bbeta}(\bx_2, y_2)\bigr| \Pi_0(d(\bx_1, y_1), d(\bx_2, y_2)),
		\end{aligned}
		\end{equation}
	where $\Pi_0$ is the optimal solution in the definition of the Wasserstein metric, i.e., it is the joint distribution of $(\bx_1, y_1)$ and $(\bx_2, y_2)$ with
	marginals $\mbb{Q}$ and $\hat{\mbb{P}}_N$ that achieves the minimum mass transportation cost. Comparing (\ref{finitedif}) with the definition of the Wasserstein distance, we wish to bound the following {\em growth rate} of $l_{\bbeta}(\bx, y)$:
	\begin{equation*}
	\frac{\bigl|l_{\bbeta}(\bx_1, y_1) - l_{\bbeta}(\bx_2, y_2)\bigr|}{s((\bx_1, y_1), (\bx_2, y_2))},  \ \forall (\bx_1, y_1), (\bx_2, y_2),
	\end{equation*}
	in order to relate $\big|\mbb{E}^{\mbb{Q}}[ l_{\bbeta}(\bx, y)] -  \mbb{E}^{\hat{\mbb{P}}_N}[ l_{\bbeta}(\bx, y)]\big|$ with $W_1 (\mbb{Q}, \ \hat{\mbb{P}}_N)$. To this end, we define a continuous and differentiable univariate function $h(a) \triangleq \log(1+\exp(-a))$, and apply the mean value theorem to it, which yields that for any $a, b\in \mbb{R}$, $\exists c \in (a,b)$ such that:
	\begin{equation*}
	\biggl|\frac{h(b) - h(a)}{b-a}\biggr| = \bigl|\bigtriangledown h(c)\bigr| =  \frac{e^{-c}}{1+e^{-c}} \le 1.
	\end{equation*}
	By noting that $l_{\bbeta}(\bx, y) = h(y\bbeta'\bx)$, we immediately have:
	\begin{equation} \label{loss-lg}
	\begin{aligned}
	\bigl|l_{\bbeta}(\bx_1, y_1) - l_{\bbeta}(\bx_2, y_2)\bigr| & \le \bigl|y_1\bbeta'\bx_1 - y_2\bbeta'\bx_2\bigr| \\
	& \le \|y_1\bx_1 - y_2 \bx_2\| \|\bbeta\|_* \\
	& \le  s((\bx_1, y_1), (\bx_2, y_2)) \|\bbeta\|_*, \ \forall (\bx_1, y_1), (\bx_2, y_2),
	\end{aligned}
	\end{equation}
	where the second step uses the Cauchy-Schwarz inequality, and the last step is due to the definition of the metric $s$ and the fact that $M$ is infinitely large. Combining (\ref{loss-lg}) with (\ref{finitedif}), it follows that for any $\mbb{Q} \in \Omega$,
	{\small \begin{equation*}
		\begin{aligned}
		\Bigl|\mbb{E}^{\mbb{Q}}\big[ l_{\bbeta}(\bx, y)\big] -  \mbb{E}^{\hat{\mbb{P}}_N}\big[ l_{\bbeta}(\bx, y)\big]\Bigr| & \le \|\bbeta\|_*\int_{\scrZ \times \scrZ} s((\bx_1, y_1), (\bx_2, y_2))  \Pi_0(d(\bx_1, y_1), d(\bx_2, y_2)) \\
		& = \|\bbeta\|_* W_1 (\mbb{Q}, \ \hat{\mbb{P}}_N) \\
		& \le \epsilon \|\bbeta\|_*.
		\end{aligned}
		\end{equation*}}
	Therefore, the DRO-LG problem can be reformulated as:
	\begin{equation*} 
	\inf\limits_{\bbeta} \mbb{E}^{\hat{\mbb{P}}_N}\big[ l_{\bbeta}(\bx, y)\big] + \epsilon \|\bbeta\|_* = \inf\limits_{\bbeta} \frac{1}{N} \sum_{i=1}^N \log\bigl(1+\exp(-y_i \bbeta'\bx_i)\bigr) + \epsilon \|\bbeta\|_*.
	\end{equation*}
	\qed
\end{proof}

\subsection*{Prediction and Estimation Performance of the GWGL-LR Estimator}
We are interested in two types of performances: (1) {\em
	Prediction quality}, or out-of-sample performance, which measures the predictive
power of the GWGL solutions on new, unseen samples. (2) {\em Estimation quality},
which measures the discrepancy between the GWGL solutions and the underlying unknown
true coefficients. 

We note that GWGL-LR is a special case of the Wasserstein DRO formulation derived in
\cite[Eq. 10]{chen2017outlier}, and thus the two types of performance guarantees
derived in \cite{chen2017outlier}, one for generalization ability (prediction error),
and the other for the discrepancy between the estimated and the true regression
coefficients (estimation error), still apply to our GWGL-LR formulation. 

We first establish a bound for the prediction bias of the solution to
the GWGL-LR formulation, where the Wasserstein metric is induced by the weighted $(2,
\infty)$-norm with weight $\bw = (\frac{1}{\sqrt{p_1}}, \ldots, \frac{1}{\sqrt{p_L}},
M)$. The dual norm in this case is just the weighted $(2, 1)$-norm with weight
$\bw^{-1}= (\sqrt{p_1}, \ldots, \sqrt{p_L}, 1/M)$. 
Throughout this section we use $\bbeta^*$ and $\hat{\bbeta}$ to denote the true and
estimated regression coefficient vectors, respectively. We first state several assumptions that are needed to establish the results.

\begin{ass} \label{a1} The weighted $(2, \infty)$-norm of the uncertainty parameter
	$(\bx, y)$ with weight $\bw = (\frac{1}{\sqrt{p_1}}, \ldots,
	\frac{1}{\sqrt{p_L}}, M)$ is bounded above by $R$ almost surely.
\end{ass}

\begin{ass} \label{a3} For every feasible $\bbeta$,
	$\|(-\bbeta^1, \ldots, -\bbeta^L, 1)_{\bw^{-1}}\|_{2, 1} \le \bar{B}$, where $\bw^{-1} = (\sqrt{p_1}, \ldots, \sqrt{p_L}, 1/M)$.
\end{ass}

Let $\hat{\bbeta}$ be an optimal solution to (\ref{gwgl-lr}),
obtained using the samples $(\bx_i, y_i)$, $i=1,\ldots,N$. Suppose we draw a
new i.i.d.\ sample $(\bx,y)$. Using Theorem~3.3 in
\cite{chen2017outlier}, Theorem~\ref{prediction2} establishes
bounds on the error $|y - \bx'\hat{\bbeta}|$.

\begin{thm} \label{prediction2} Under Assumptions \ref{a1} and
	\ref{a3}, for any $0<\delta<1$, with probability at least $1-\delta$
	with respect to the sampling,
	\begin{equation*}
	\mathbb{E}[|y - \bx'\hat{\bbeta}|]\le
	\frac{1}{N}\sum_{i=1}^N
	|y_i - \bx_i'\hat{\bbeta}|+\frac{2\bar{B}R}{\sqrt{N}}+ 
	\bar{B}R\sqrt{\frac{8\log(2/\delta)}{N}}\ ,
	\end{equation*}
	and for any $\zeta>(2\bar{B}R/\sqrt{N})+
	\bar{B}R\sqrt{8\log(2/\delta)/N}$,
	\begin{equation*}
	\mathbb{P}\biggl(|y - \bx'\hat{\bbeta}| \ge
	\frac{1}{N}\sum_{i=1}^N |y_i - \bx_i'\hat{\bbeta}|+\zeta\biggr) 
	\le \frac{\frac{1}{N}\sum_{i=1}^N
		|y_i - \bx_i'\hat{\bbeta}|+\frac{2\bar{B}R}{\sqrt{N}}+ 
		\bar{B}R\sqrt{\frac{8\log(2/\delta)}{N}}}{\frac{1}{N}\sum_{i=1}^N
		|y_i - \bx_i'\hat{\bbeta}|+\zeta}. 
	\end{equation*}
\end{thm}

Theorem~\ref{prediction2} essentially says that with a high probability, the expected
loss on new test samples using our GWGL-LR estimator can be upper bounded by the
average loss in the training samples plus two terms that are related to the magnitude
of the regularizer $\bar{B}$, the uncertainty level $R$, the confidence level
$\delta$, and converge to zero as $O(1/\sqrt{N})$. This result justifies the form of
the regularizer used in (\ref{gwgl-lr}) and guarantees a small generalization error
of the GWGL-LR solution.

We next discuss the estimation performance of the GWGL-LR solution.
Theorem \ref{estimation2}, a specialization of Theorem 3.11 in \cite{chen2017outlier},
provides a bound for the estimation bias in the GWGL-LR formulation. We first state the assumptions that are needed to establish the result.

\begin{ass} \label{2norm} The $\ell_2$ norm of $(-\bbeta, 1)$
	is bounded above by $\bar{B}_2$.
\end{ass}

\begin{ass} \label{RE} For some
	set $$\scrA(\bbeta^*) \coloneqq \text{cone}\{\bv|\
	\|(-\bbeta^*, 1)_{\bw^{-1}}+\bv_{\bw^{-1}}\|_{2,
		1}\le\|(-\bbeta^*, 1)_{\bw^{-1}}\|_{2, 1}\} 
	\cap \mathbb{S}^{p+1}$$ and some positive scalar $\underline{\alpha}$, the following holds,
	\begin{equation*}
	\inf\limits_{\bv\in \scrA(\bbeta^*)}\bv'\bZ\bZ'\bv\ge\underline{\alpha},
	\end{equation*}
	where $\bZ=[(\bx_1, y_1), \ldots, (\bx_N, y_N)]$ is the matrix with columns $(\bx_i, y_i), 
	i = 1, \ldots, N$, and $\mathbb{S}^{p+1}$ is the unit sphere in the $(p+1)$-dimensional Euclidean space.
\end{ass}


\begin{ass} \label{subgaussian} $(\bx, y)$ is a centered sub-Gaussian
	random vector, i.e., it has zero mean and satisfies the following
	condition:
	\begin{equation*}
	\vertiii{(\bx, y)}_{\psi_2}=\sup\limits_{\bu\in
		\mathbb{S}^{p+1}}\vertiii{(\bx, y)'\bu}_{\psi_2}\le \mu.
	\end{equation*}
\end{ass}

\begin{ass} \label{eigenv}
	The covariance matrix of $(\bx, y)$ has bounded positive
	eigenvalues. Set $\bGamma=\mathbb{E}[(\bx, y)(\bx, y)']$; then, 
	\begin{equation*}
	0<\lambda_{\text{min}} \triangleq
	\lambda_{\text{min}}(\bGamma)\le\lambda_{\text{max}}(\bGamma)\triangleq
	\lambda_{\text{max}}<\infty. 
	\end{equation*}
\end{ass}

\begin{defi}[Sub-Gaussian random variable]
	A random variable $z$ is sub-Gaussian if it is zero mean, and the $\psi_2$-norm defined
	below is finite, i.e.,
	\begin{equation*}
	\vertiii{z}_{\psi_2}\triangleq \sup_{q \ge 1}
	\frac{(\mbb{E}|z|^q)^{1/q}}{\sqrt{q}} < +\infty. 
	\end{equation*}
\end{defi}
An equivalent property for sub-Gaussian random variables is that their
tail distribution decays at least as fast as a Gaussian, namely, 
\[ 
\mbb{P}(|z|\geq t) \leq 2 \exp\{-t^2/C^2\},\quad \forall t \geq 0, 
\] 
for some constant $C$. A random vector $\bz \in \mbb{R}^{p+1}$ is sub-Gaussian if $\bz'\bu$ is
sub-Gaussian for any $\bu \in \mbb{R}^{p+1}$. The $\psi_2$-norm of a vector
$\bz$ is defined as: 
\[
\vertiii{\bz}_{\psi_2} \triangleq \sup\limits_{\bu\in
	\mbb{S}^{p+1}}\vertiii{\bz'\bu}_{\psi_2},
\] 
where $\mbb{S}^{p+1}$ denotes the unit sphere in the $(p+1)$-dimensional
Euclidean space.

\begin{defi}[Gaussian width]
	For any set $\scrA \subseteq \mbb{R}^{p+1}$, its Gaussian width is
	defined as:
	\begin{equation*}
	w(\scrA) \triangleq \mbb{E}\Bigl[\sup_{\bu \in \scrA} \bu'\bg\Bigr],
	\end{equation*}
	where $\bg\sim {\cal N}(\bzero,\bI)$ is a $(p+1)$-dimensional
	standard Gaussian random vector.
\end{defi}

\begin{thm} \label{estimation2} 
	Suppose the true regression coefficient vector is $\bbeta^*$
	and the solution to GWGL-LR is
	$\hat{\bbeta}$. Under Assumptions \ref{a1}, \ref{2norm},
	\ref{RE}, \ref{subgaussian}, and \ref{eigenv}, when the sample size $N\ge
	\bar{C_1}\bar{\mu}^4 \mu_0^2
	(\lambda_{\text{max}}/\lambda_{\text{min}})\cdot 
	(w(\scrA(\bbeta^*))+3)^2$, with probability at least \\
	$1-\exp(-C_2N/\bar{\mu}^4)-C_4\exp(-C_5^2 (w(\scrB_u))^2/(4\rho^2))$,	
	\begin{equation*}
	\|\hat{\bbeta}-\bbeta^*\|_2\le
	\frac{\bar{C}R\bar{B}_2\mu}{N\lambda_{\text{min}}}
	w(\scrB_u)\Psi(\bbeta^*),
	\end{equation*}	
	where $\bar{\mu}=\mu\sqrt{(1/\lambda_{\text{min}})}$; $\mu_0$ is
	the $\psi_2$-norm of a standard Gaussian random vector $\bg \in
	\mathbb{R}^{p+1}$; $w(\scrA(\bbeta^*))$ is the Gaussian width (defined below)
	of $\scrA(\bbeta^*)$ (cf. Assumption \ref{RE}); $w(\scrB_u)$ is the Gaussian width of $\scrB_u$,
	where $\scrB_u$ is the unit ball of the norm $\|\cdot\|_\infty$;
	$\rho=\sup_{\bv\in \scrB_u}\|\bv\|_2$;
	$\Psi(\bbeta^*)=\sup_{\bv\in
		\scrA(\bbeta^*)}\|\bv_{\bw^{-1}}\|_{2, 1}$; and $\bar{C_1}, C_2, C_4, C_5,
	\bar{C}$ are positive constants.
\end{thm}

With Theorem \ref{estimation2}, we are able to provide bounds for some popular
performance metrics, such as the {\em Relative Risk (RR)}, {\em Relative
	Test Error (RTE)}, and {\em Proportion of Variance Explained (PVE)}
\cite{hastie2017extended}. All these metrics evaluate the accuracy of
the regression coefficient estimates on a new test sample drawn from the
same probability distribution as the training samples. Let $(\bx_0,
y_0)$ be such a test sample satisfying
$y_0 = \bx_0' \bbeta^* + \eta_0$,
where $\eta_0$ is a random noise with zero mean and variance $\sigma^2$, and is independent of the zero mean predictor $\bx_0$.  For a fixed set of
training samples, let the solution to GWGL-LR be
$\hat{\bbeta}$. As in
\cite{hastie2017extended}, define
\begin{equation*}
\text{RR}(\hat{\bbeta}) = \frac{\mbb{E}(\bx_0'\hat{\bbeta} - \bx_0'
	\bbeta^*)^2}{\mbb{E}(\bx_0' \bbeta^*)^2} = \frac{(\hat{\bbeta} -
	\bbeta^*)'\mathbf{\Sigma}(\hat{\bbeta} - \bbeta^*)}{(\bbeta^*)'
	\mathbf{\Sigma} \bbeta^*}, 
\end{equation*}
where $\mathbf{\Sigma}$ is the covariance matrix of $\bx_0$, which is
just the top left block of the matrix $\bGamma$ in
Assumption~\ref{eigenv}. 
RTE is defined as:
\begin{equation*}
\text{RTE}(\hat{\bbeta}) = \frac{\mbb{E}(y_0 - \bx_0'\hat{\bbeta})^2}{\sigma^2}
= \frac{(\hat{\bbeta} - \bbeta^*)'\bSigma (\hat{\bbeta} - \bbeta^*) +
	\sigma^2}{\sigma^2}. 
\end{equation*}
PVE is defined as:
\begin{equation*}
\text{PVE}(\hat{\bbeta}) = 1- \frac{\mbb{E}(y_0 - \bx_0'
	\hat{\bbeta})^2}{Var(y_0)} = 1 - \frac{(\hat{\bbeta} -
	\bbeta^*)'\bSigma (\hat{\bbeta} - \bbeta^*) + \sigma^2}{(\bbeta^*)'
	\mathbf{\Sigma} \bbeta^* + \sigma^2}. 
\end{equation*}
Using Theorem \ref{estimation2}, we can bound the term $(\hat{\bbeta}
- \bbeta^*)'\bSigma (\hat{\bbeta} - \bbeta^*)$ as follows:
\begin{equation} \label{metricbound}
(\hat{\bbeta} - \bbeta^*)'\bSigma (\hat{\bbeta} - \bbeta^*) \le \lambda_{max}(\bSigma) \|\hat{\bbeta} - \bbeta^*\|_2^2 \le \lambda_{max}(\bSigma) \biggl(\frac{\bar{C}R\bar{B}_2\mu}{N\lambda_{\text{min}}}
w(\scrB_u)\Psi(\bbeta^*)\biggr)^2,
\end{equation}
where $\lambda_{max}(\bSigma)$ is the maximum eigenvalue of
$\bSigma$. Using (\ref{metricbound}), bounds for RR, RTE, and PVE can be
readily obtained and are summarized in the following Corollary.

\begin{col} \label{estimation-1rte} Under the specifications in
	Theorem~\ref{estimation2}, when the sample size $$N\ge
	\bar{C_1}\bar{\mu}^4 \mu_0^2
	(\lambda_{\text{max}}/\lambda_{\text{min}})
	(w(\scrA(\bbeta^*))+3)^2,$$ with probability at least 
	$1-\exp(-C_2N/\bar{\mu}^4)- C_4\exp(-C_5^2 (w(\scrB_u))^2/(4\rho^2))$,
	\begin{equation*}
	\text{RR}(\hat{\bbeta}) \le \frac{\lambda_{max}(\bSigma)
		\biggl(\frac{\bar{C}R\bar{B}_2\mu}{N\lambda_{\text{min}}} 
		w(\scrB_u)\Psi(\bbeta^*)\biggr)^2}{(\bbeta^*)'
		\mathbf{\Sigma} \bbeta^*}, 
	\end{equation*} 
	\begin{equation*}
	\text{RTE}(\hat{\bbeta}) \le \frac{ \lambda_{max}(\bSigma)
		\biggl(\frac{\bar{C}R\bar{B}_2\mu}{N\lambda_{\text{min}}} 
		w(\scrB_u)\Psi(\bbeta^*)\biggr)^2 + \sigma^2}{\sigma^2},
	\end{equation*}
	and,
	\begin{equation*}
	\text{PVE}(\hat{\bbeta}) \ge 1 - \frac{\lambda_{max}(\bSigma)
		\biggl(\frac{\bar{C}R\bar{B}_2\mu}{N\lambda_{\text{min}}} 
		w(\scrB_u)\Psi(\bbeta^*)\biggr)^2 + \sigma^2}{(\bbeta^*)'
		\mathbf{\Sigma} \bbeta^* + \sigma^2}, 
	\end{equation*}
	where all parameters are defined in the same way as in
	Theorem~\ref{estimation2}.
\end{col}

\subsection*{Predictive Performance of the GWGL-LG Estimator}
In this subsection we establish bounds on the prediction error of the GWGL-LG
solution. Similar to \cite{chen2017outlier}, we
will use the {\em Rademacher complexity} of the class of logloss (negative
log-likelihood) functions to bound the generalization error. Two assumptions that
impose conditions on the magnitude of the regularizer and the uncertainty level of
the predictor are needed.

\begin{ass} \label{a1-lg} The weighted $(2, \infty)$-norm of $\bx$ with weight $\bw = (\frac{1}{\sqrt{p_1}}, \ldots, \frac{1}{\sqrt{p_L}})$ is
	bounded above almost surely, i.e., $\|\bx_{\bw}\|_{2, \infty} \le R_{\bx}$.
\end{ass}

\begin{ass} \label{a2-lg} The weighted $(2, 1)$-norm of $\bbeta$ with $\bw^{-1} = (\sqrt{p_1}, \ldots, \sqrt{p_L})$ is bounded
	above, namely,
	$\sup_{\bbeta}\|\bbeta_{\bw^{-1}}\|_{2, 1}=\bar{B}_1$.
	
\end{ass}

Under these two assumptions, the logloss could be bounded via the Cauchy-Schwarz inequality.
\begin{lem} \label{l1}
	Under Assumptions~\ref{a1-lg} and \ref{a2-lg}, it follows
	\begin{equation*}
	\log\big(1+\exp(-y \bbeta'\bx)\big) \le \log\big(1 + \exp(R_{\bx} \bar{B}_1)\big), \quad \text{almost surely}.
	\end{equation*}
\end{lem} 

Now consider the following class of loss functions: 
\begin{equation*}
\scrL=\big\{(\bx, y) \mapsto l_{\bbeta}(\bx, y): l_{\bbeta}(\bx, y)= \log\big(1+\exp(-y \bbeta'\bx)\big),\
\|\bbeta_{\bw^{-1}}\|_{2, 1}\le \bar{B}_1 \big\}. 
\end{equation*}
It follows from \cite{chen2017outlier,Dim14} that the empirical {\em Rademacher complexity} of $\scrL$, denoted by $\scrR_N(\scrL)$, can be upper bounded by:
\begin{equation*}
\scrR_N(\scrL)\le \frac{2 \log\big(1 + \exp(R_{\bx} \bar{B}_1)\big)}{\sqrt{N}}.
\end{equation*}
Then, applying Theorem 8 in \cite{Peter02}, we have the following result on the prediction error of our GWGL-LG estimator.

\begin{thm} \label{prediction-lg} Let $\hat{\bbeta}$ be an optimal solution to (\ref{gwgl-lg}), obtained using
	$N$ training samples $(\bx_i, y_i)$, $i=1,\ldots,N$. Suppose we draw a new i.i.d.\
	sample $(\bx,y)$. Under Assumptions \ref{a1-lg} and \ref{a2-lg}, for any
	$0<\delta<1$, with probability at least $1-\delta$ with respect to the
	sampling,
	{\small \begin{equation} \label{exp}
		\begin{aligned}
		\mathbb{E}\big[\log\big(1+\exp(-y \bx'\hat{\bbeta})\big)\big] & \le
		\frac{1}{N}\sum_{i=1}^N
		\log\big(1+\exp(-y_i \bx_i'\hat{\bbeta})\big)+\frac{2 \log\big(1 + \exp(R_{\bx} \bar{B}_1)\big)}{\sqrt{N}} \\ 
		& \quad + 
		\log\big(1 + \exp(R_{\bx} \bar{B}_1)\big)\sqrt{\frac{8\log(2/\delta)}{N}}\ ,
		\end{aligned}
		\end{equation}}
	and for any $\zeta>\frac{2 \log(1 + \exp(R_{\bx} \bar{B}_1))}{\sqrt{N}} 
	+ 
	\log\big(1 + \exp(R_{\bx} \bar{B}_1)\big)\sqrt{\frac{8\log(2/\delta)}{N}}$,
	{\footnotesize \begin{equation} \label{prob}
		\begin{aligned}
		& \ \mathbb{P}\Bigl( \log\big(1+\exp( -y \bx'\hat{\bbeta})\big)  \ge
		\frac{1}{N}\sum_{i=1}^N \log\big(1+\exp(-y_i \bx_i'\hat{\bbeta})\big)+\zeta \Bigr) 
		\\ \le & \ 
		\frac{\frac{1}{N}\sum_{i=1}^N
			\log\big(1+\exp(-y_i \bx_i'\hat{\bbeta})\big)+\frac{2 \log(1 + \exp(R_{\bx} \bar{B}_1))}{\sqrt{N}} 
			+ 
			\log\big(1 + \exp(R_{\bx} \bar{B}_1)\big)\sqrt{\frac{8\log(2/\delta)}{N}}}{\frac{1}{N}\sum_{i=1}^N
			\log\big(1+\exp(-y_i \bx_i'\hat{\bbeta})\big)+\zeta}\ . 
		\end{aligned}
		\end{equation}}
\end{thm}

Theorem \ref{prediction-lg} implies that the groupwise regularized LG formulation
(\ref{gwgl-lg}) yields a solution with a small generalization error on new
i.i.d. samples. 

\subsection*{Proof of Theorem \ref{grouping2} for GWGL-LR}
\begin{proof} By the optimality condition associated with formulation (\ref{gwgl-lr}),
	$\hat{\bbeta}$ satisfies:
	\begin{equation} \label{opt3} \bx_{,i}'\text{sgn}(\by-\bX \hat{\bbeta})
	= N \epsilon \sqrt{p_{l_1}}
	\frac{\hat{\beta}_i}{\|\hat{\bbeta}^{l_1}\|_2} ,
	\end{equation}
	\begin{equation} \label{opt4} \bx_{,j}'\text{sgn}(\by-\bX \hat{\bbeta})
	= N \epsilon \sqrt{p_{l_2}}
	\frac{\hat{\beta}_j}{\|\hat{\bbeta}^{l_2}\|_2},
	\end{equation}
	where the $\text{sgn}(\cdot)$ function is applied to a vector elementwise. Subtracting (\ref{opt4}) from (\ref{opt3}), we obtain:
	\begin{equation*}
	(\bx_{,i}- \bx_{,j})'\text{sgn}(\by-\bX \hat{\bbeta}) = N \epsilon
	\Biggl( \frac{\sqrt{p_{l_1}}\hat{\beta}_i}{\|\hat{\bbeta}^{l_1}\|_2} -
	\frac{\sqrt{p_{l_2}}\hat{\beta}_j}{\|\hat{\bbeta}^{l_2}\|_2}\Biggr). 
	\end{equation*} 
	Using the Cauchy-Schwarz inequality and $\|\bx_{,i}-
	\bx_{,j}\|_2^2=2(1-\rho)$, we obtain
	\begin{equation*}
	\begin{split}
	D(i, j) & = \Biggl|\frac{\sqrt{p_{l_1}}\hat{\beta}_i}{\|\hat{\bbeta}^{l_1}\|_2} -  \frac{\sqrt{p_{l_2}}\hat{\beta}_j}{\|\hat{\bbeta}^{l_2}\|_2}\Biggr| \\
	& \le  \frac{1}{N \epsilon} \|\bx_{,i}- \bx_{,j}\|_2
	\|\text{sgn}(\by-\bX \hat{\bbeta})\|_2 \\ 
	& \le \frac{\sqrt{2(1-\rho)}}{\sqrt{N}\epsilon}.  
	\end{split}
	\end{equation*} 
	\qed
\end{proof}

\subsection*{Proof of Theorem \ref{grouping2} for GWGL-LG}

\begin{proof} By the optimality condition associated with formulation (\ref{gwgl-lg}),
	$\hat{\bbeta}$ satisfies:
	\begin{equation} \label{opt3-lg} 
	\sum_{k=1}^N \frac{\exp(-y_k \bx_k' \hat{\bbeta})}{1+ \exp(-y_k \bx_k' \hat{\bbeta})} y_k x_{k,i}
	= N \epsilon \sqrt{p_{l_1}}
	\frac{\hat{\beta}_i}{\|\hat{\bbeta}^{l_1}\|_2} ,
	\end{equation}
	\begin{equation} \label{opt4-lg} 
	\sum_{k=1}^N \frac{\exp(-y_k \bx_k' \hat{\bbeta})}{1+ \exp(-y_k \bx_k' \hat{\bbeta})} y_k x_{k,j}
	= N \epsilon \sqrt{p_{l_2}}
	\frac{\hat{\beta}_j}{\|\hat{\bbeta}^{l_2}\|_2},
	\end{equation}
	where $x_{k,i}$ and $x_{k, j}$ denote the $i$-th and $j$-th elements of $\bx_k$, respectively.
	Subtracting (\ref{opt4-lg}) from (\ref{opt3-lg}), we get:
	\begin{equation} \label{derdiff}
	\sum_{k=1}^N \frac{\exp(-y_k \bx_k' \hat{\bbeta})}{1+ \exp(-y_k \bx_k' \hat{\bbeta})} \big(y_k x_{k,i} - y_k x_{k,j}\big)
	= N \epsilon
	\Biggl( \frac{\sqrt{p_{l_1}}\hat{\beta}_i}{\|\hat{\bbeta}^{l_1}\|_2} -
	\frac{\sqrt{p_{l_2}}\hat{\beta}_j}{\|\hat{\bbeta}^{l_2}\|_2}\Biggr). 
	\end{equation} 
	Note that the LHS of~\ref{derdiff} can be written as $\bv_1'\bv_2$, where 
	$$\bv_1 = \bigg( \frac{\exp(-y_1 \bx_1' \hat{\bbeta})}{1+ \exp(-y_1 \bx_1' \hat{\bbeta})}, \ldots, \frac{\exp(-y_N \bx_N' \hat{\bbeta})}{1+ \exp(-y_N \bx_N' \hat{\bbeta})}\bigg),$$ $$\bv_2 = \big( y_1 ( x_{1,i} - x_{1,j}), \ldots, y_N (x_{N,i} - x_{N,j})\big).$$
	Using the Cauchy-Schwarz inequality and $\|\bx_{,i}-
	\bx_{,j}\|_2^2=2(1-\rho)$, we obtain
	\begin{equation*}
	\begin{split}
	D(i, j) & = \Biggl|\frac{\sqrt{p_{l_1}}\hat{\beta}_i}{\|\hat{\bbeta}^{l_1}\|_2} -  \frac{\sqrt{p_{l_2}}\hat{\beta}_j}{\|\hat{\bbeta}^{l_2}\|_2}\Biggr| \\
	& \le \frac{1}{N \epsilon} \|\bv_1\|_2 \|\bv_2\|_2 \\
	& \le  \frac{1}{N \epsilon} \sqrt{N} \|\bx_{,i}- \bx_{,j}\|_2
	\\ 
	& = \frac{\sqrt{2(1-\rho)}}{\sqrt{N}\epsilon}.  
	\end{split}
	\end{equation*} 
	\qed
\end{proof}

\section*{Appendix B: Omitted Numerical Results}
This section contains the experimental setup and results that are omitted in Section \ref{s4}.

\subsection*{Omitted Results in Section \ref{gwgl-lr-exp}}
\subsubsection*{Hyperparameter Tuning}
All the penalty parameters are tuned using a separate validation
dataset. Specifically, we divide all the $N$ training samples into two sets, dataset
1 and dataset 2 (validation set). For a pre-specified range of values for the penalty
parameters, dataset 1 is used to train the models and derive $\hat{\bbeta}$, and the
performance of $\hat{\bbeta}$ is evaluated on dataset 2. We choose the penalty
parameter that yields the minimum unpenalized loss of the respective approaches on
the validation set. As to the range of values for the tuned parameters, we borrow
ideas from \cite{hastie2017extended}, where the LASSO was tuned over $50$ values
ranging from $\lambda_m \triangleq \|\bX'\by\|_{\infty}$ to a small fraction of
$\lambda_m$ on a log scale. In our experiments, this range is properly adjusted for
the GLASSO estimators. Specifically, for GWGL and GSRL, the tuning range is:
$\sqrt{\exp(\text{lin}(\log(0.005\cdot
	\|\bX'\by\|_{\infty}),\log(\|\bX'\by\|_{\infty}),50)) /\max(p_1, \ldots, p_L)},$
where the \\ function $\text{lin}(a, b, n)$ takes in scalars $a$, $b$ and $n$
(integer) and outputs a set of $n$ values equally spaced between $a$ and $b$; the
$\exp$ function is applied elementwise to a vector. Compared to LASSO, the values are
scaled by $\max(p_1, \ldots, p_L)$, and the square root operation is due to the
$\ell_1$-loss function, or the square root of the $\ell_2$-loss used in these
formulations. For the GLASSO with $\ell_2$-loss, the range is:
$\exp(\text{lin}(\log(0.005\cdot
\|\bX'\by\|_{\infty}),\log(\|\bX'\by\|_{\infty}),50)) /\sqrt{\max(p_1, \ldots,
	p_L)}.$

\subsubsection*{Implementation of Spectral Clustering}
In
our implementation, the $k$-nearest neighbor similarity graph is constructed, where
we connect $\bx_{,i}$ and $\bx_{,j}$ with an undirected edge if $\bx_{,i}$ is among
the $k$-nearest neighbors of $\bx_{,j}$ (in the sense of Euclidean distance) {\em or}
if $\bx_{,j}$ is among the $k$-nearest neighbors of $\bx_{,i}$. The parameter $k$ is
chosen such that the resulting graph is connected. Recall that we use the Gaussian
similarity function
\begin{equation}\label{gs}
\text{Gs}(\bx_{,i}, \bx_{,j}) \triangleq \exp\big( -\|\bx_{,i} - \bx_{,j}\|_2^2/(2\sigma_s^2)\big),
\end{equation}
to construct the graph. The scale parameter $\sigma_s$ in
(\ref{gs}) is set to the mean distance of a point to its $k$-th nearest neighbor
\cite{von2007tutorial}.  We assume that the number of clusters is known in order to
perform spectral clustering, but in case it is unknown, the eigengap heuristic
\cite{von2007tutorial} can be used, where the goal is to choose the number of
clusters $c$ such that all eigenvalues $\lambda_1, \ldots, \lambda_c$ of the graph
Laplacian are very small, but $\lambda_{c+1}$ is relatively large.

\subsubsection*{The MPI Values for GWGL-LR}
Recall that we define the {\em Maximum Percentage Improvement (MPI)} to be the maximum percentage difference of the performance metrics between GWGL-LR and
the best among all others. In Tables~\ref{table1} and \ref{table2}
we summarize the MPI brought about by our
methods compared to other procedures, when varying the SNR and $\rho_w$,
respectively. In all tables, the number outside the parentheses is the MPI value
corresponding to each metric, while the number in the parentheses indicates the value
of SNR/$\rho_w$ at which the MPI is attained.

\begin{table}[ht]
	\caption{MPI of all metrics when varying
		the SNR.} \label{table1} 
	\begin{center}
		{\small \begin{tabular}{ c|>{\centering\arraybackslash}p{2.2cm}|>{\centering\arraybackslash}p{2.2cm}|>{\centering\arraybackslash}p{2.2cm}|>{\centering\arraybackslash}p{2.2cm} } 
				\hline
				& MAD & RR   & RTE   & PVE \\ \hline 
				$q = 20\%$ & 13.7 (0.5) & 41.4 (1.47) & 13.1 (1.47) & 68.9 (0.79)  \\ 
				$q = 30\%$ & 14.7 (1.08) & 40.9 (1.08) &  17 (1.08) & 85.7 (0.68)  \\ 
				\hline
			\end{tabular}}
		\end{center}
	\end{table}
	
\begin{table}[ht]
	\caption{MPI of all metrics when varying the within group correlation.} \label{table2} 
	\begin{center}
		{\small \begin{tabular}{ c|>{\centering\arraybackslash}p{2.2cm}|>{\centering\arraybackslash}p{2.2cm}|>{\centering\arraybackslash}p{2.2cm}|>{\centering\arraybackslash}p{2.2cm} } 
				\hline
				& MAD & RR   & RTE   & PVE \\ \hline 
				$q = 20\%$ & 8.2 (0.1) & 80.5 (0.9) & 31.8 (0.9)  & 145.4 (0.9)  \\ 
				$q = 30\%$ & 10.2 (0.1) & 41.9 (0.1) & 16.7 (0.1) & 162.5 (0.1) \\ 
				\hline
			\end{tabular}}
	\end{center}
\end{table} 

\subsubsection*{The Impact of SNR and $\rho_w$ on the Performance Metrics when $q = 20\%$}
\begin{figure}[ht] 
	\begin{subfigure}{.49\textwidth}
		\centering
		\includegraphics[width=1\textwidth]{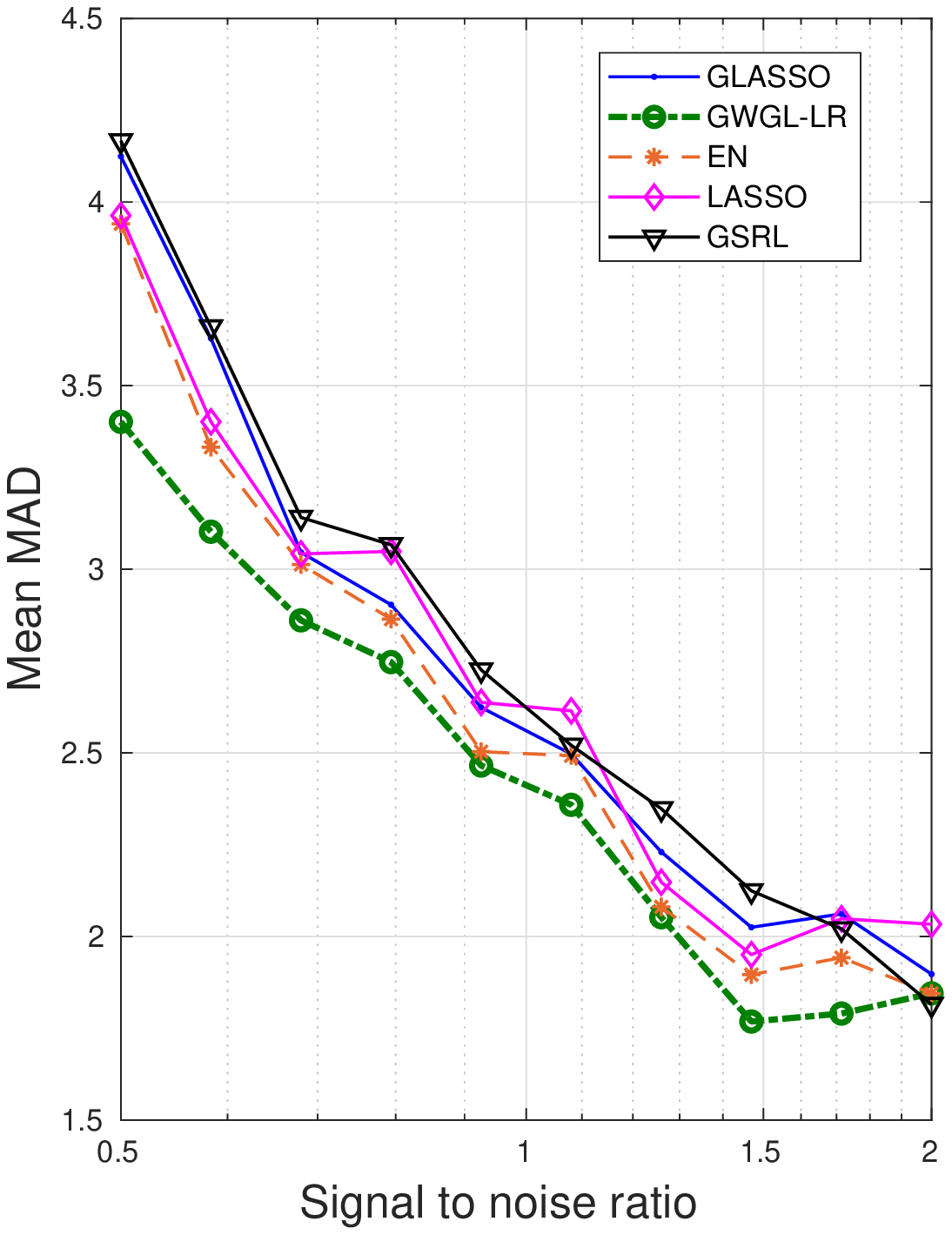}
		\caption{\small{Median Absolute Deviation.}}
	\end{subfigure}
	\begin{subfigure}{0.49\textwidth}
		\centering
		\includegraphics[width=1\textwidth]{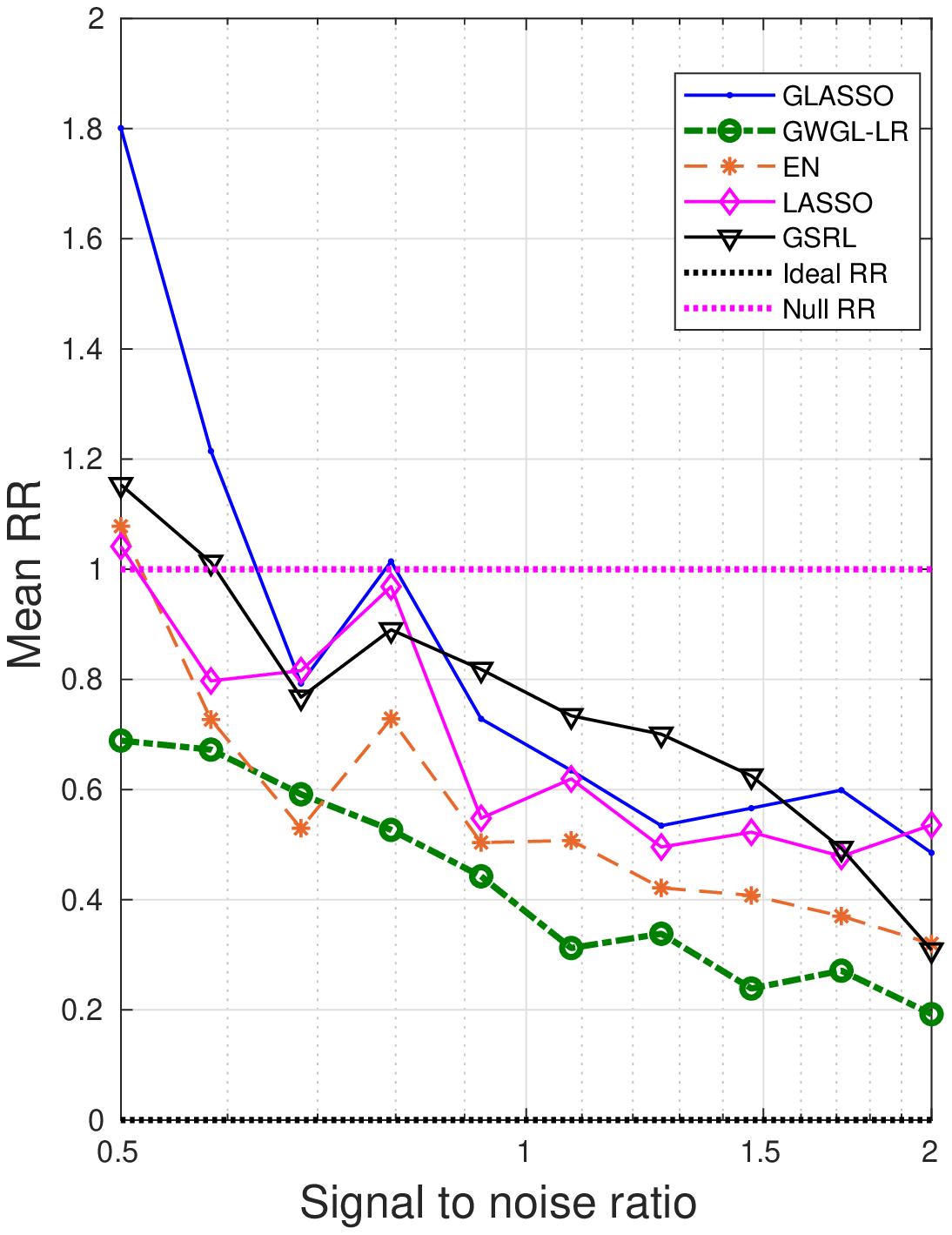}
		\caption{\small{Relative risk.}}
	\end{subfigure}
	
	\begin{subfigure}{0.49\textwidth}
		\centering
		\includegraphics[width=1\textwidth]{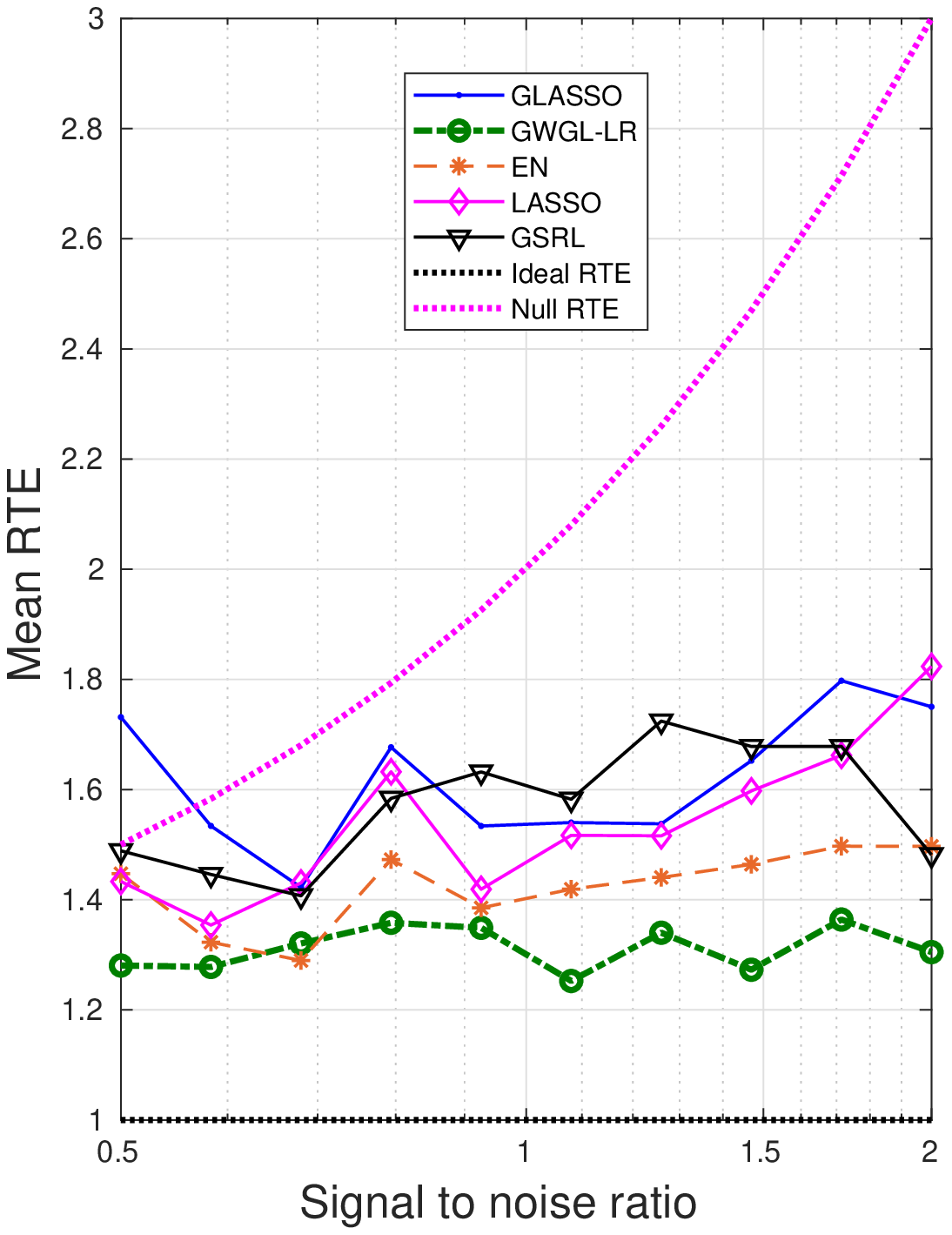}
		\caption{\small{Relative test error.}}
	\end{subfigure}%
	\begin{subfigure}{0.49\textwidth}
		\centering
		\includegraphics[width=1\textwidth]{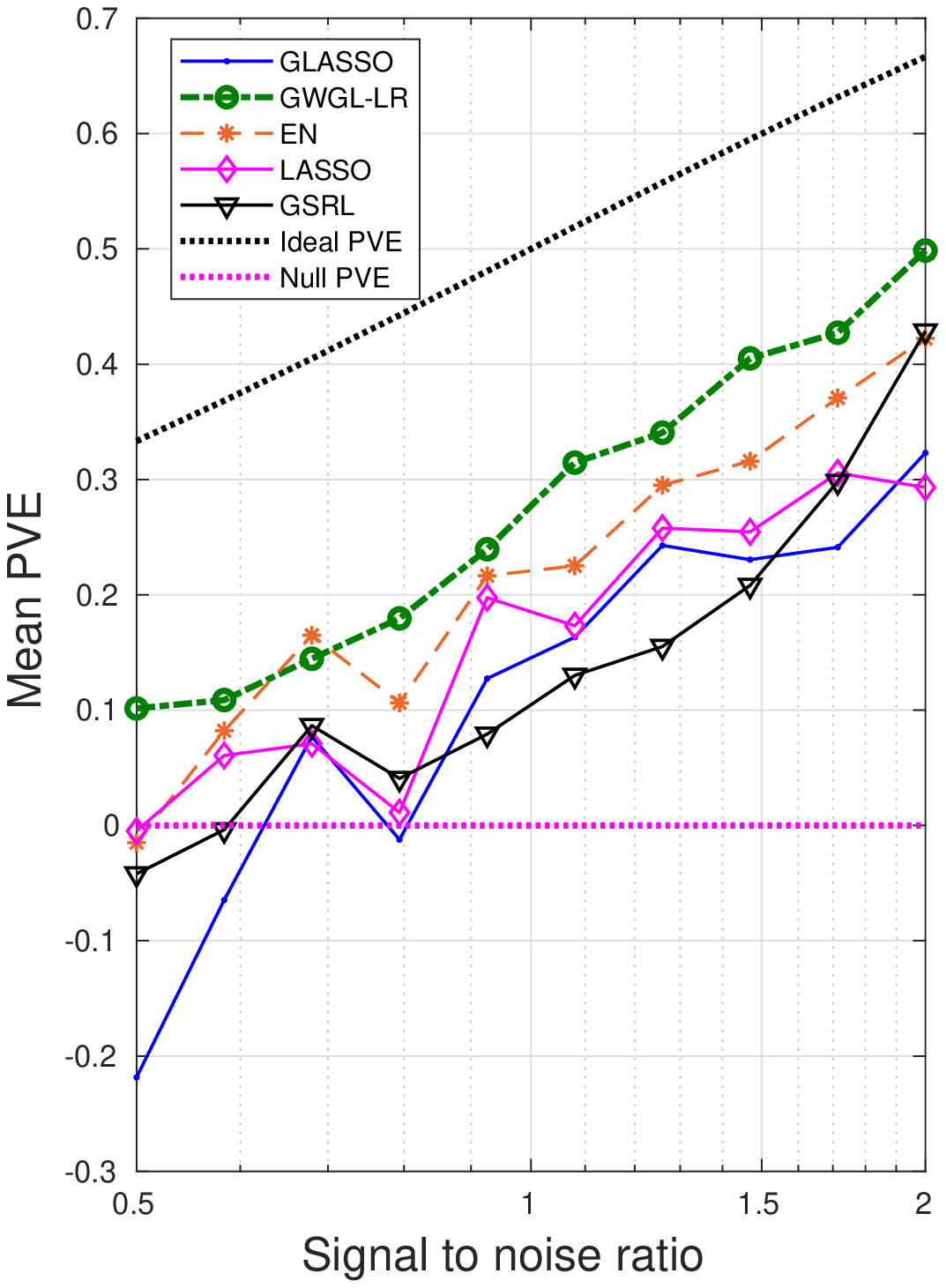}
		\caption{\small{Proportion of variance explained.}}
	\end{subfigure}
	\caption{The impact of SNR on the performance metrics, $q =
		20\%$.} \label{snr-20}
\end{figure}

\begin{figure}[h] 
	\begin{subfigure}{.49\textwidth}
		\centering
		\includegraphics[width=1\textwidth]{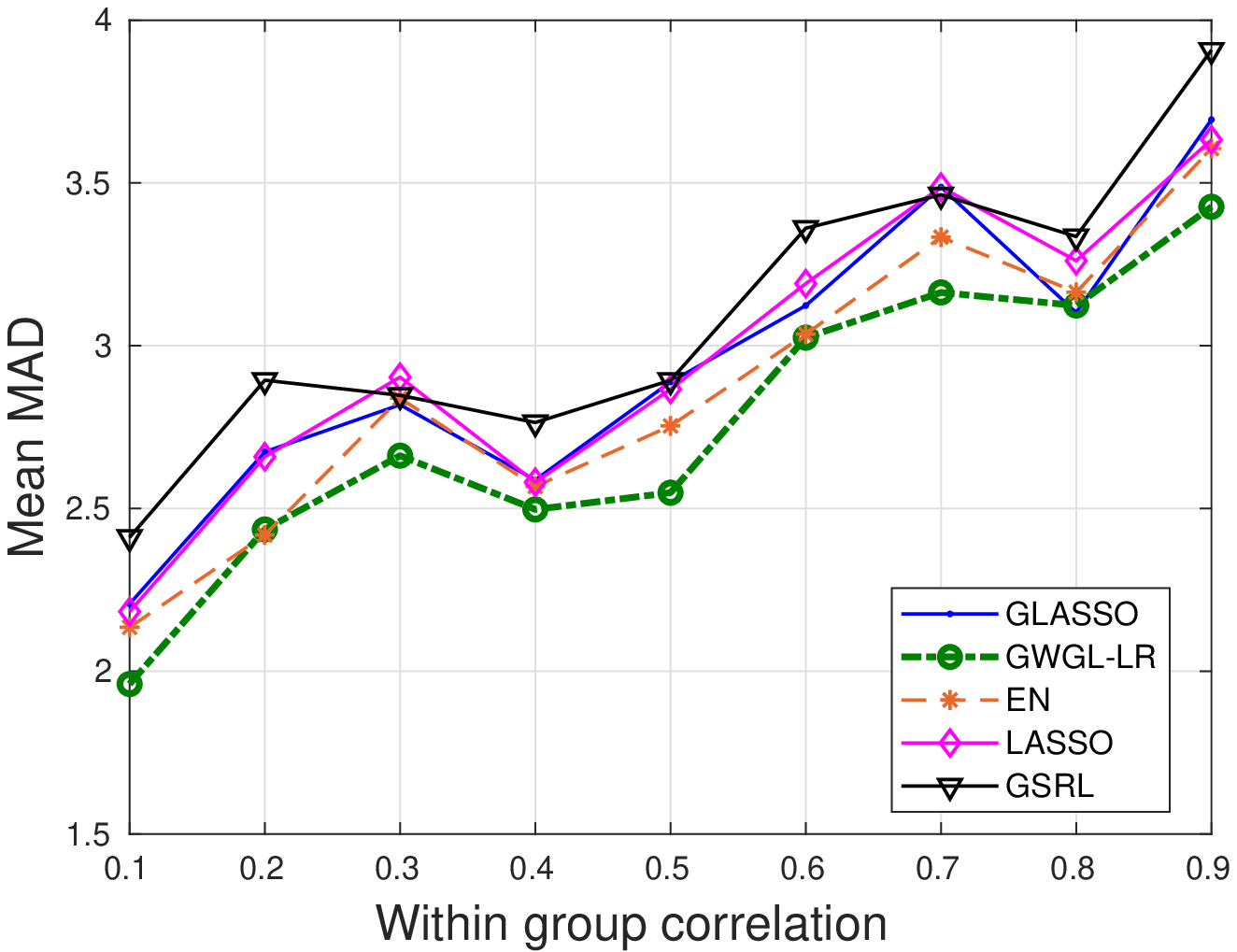}
		\caption{\small{Median Absolute Deviation.}}
	\end{subfigure}
	\begin{subfigure}{0.49\textwidth}
		\centering
		\includegraphics[width=1\textwidth]{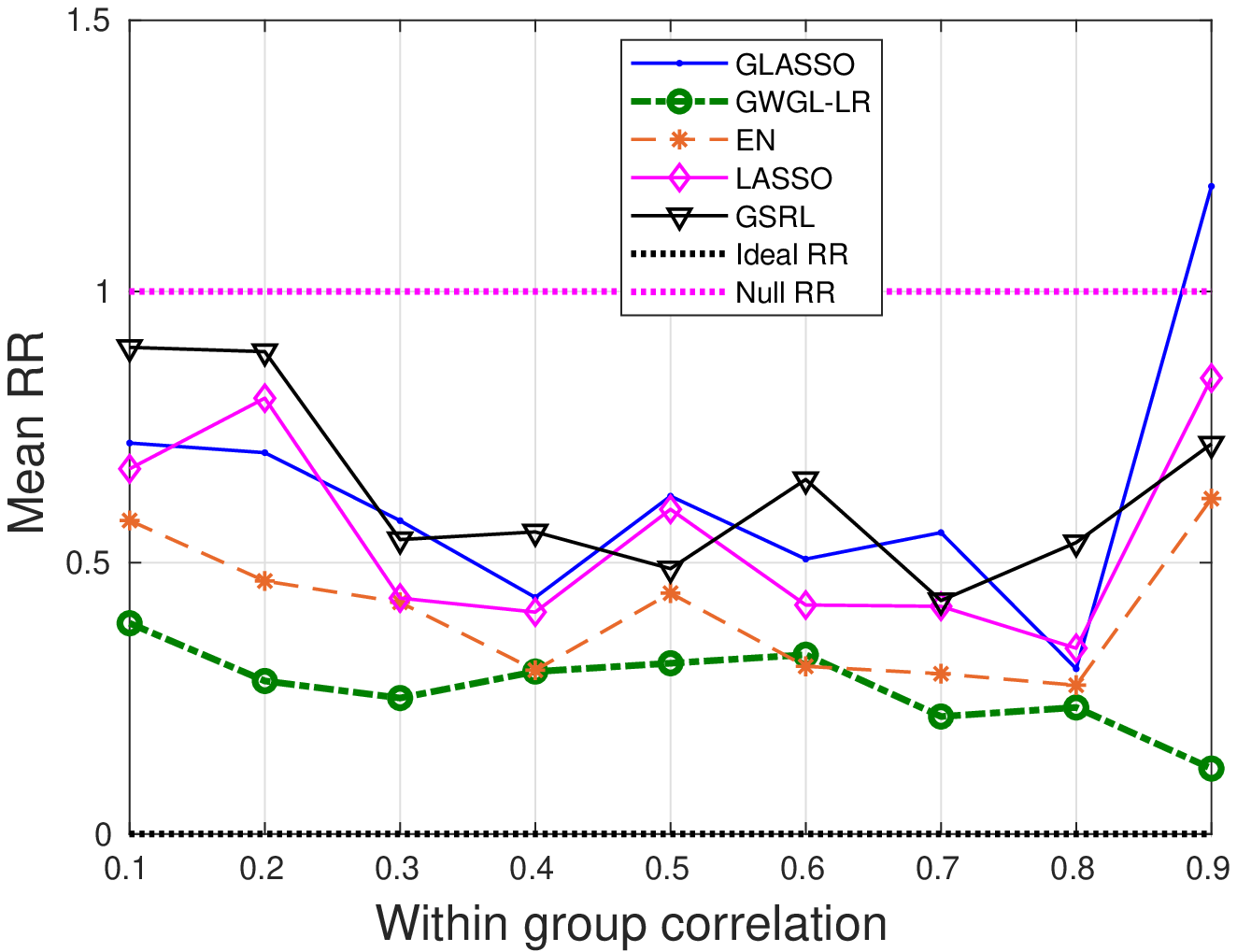}
		\caption{\small{Relative risk.}}
	\end{subfigure}
	
	\begin{subfigure}{0.49\textwidth}
		\centering
		\includegraphics[width=1\textwidth]{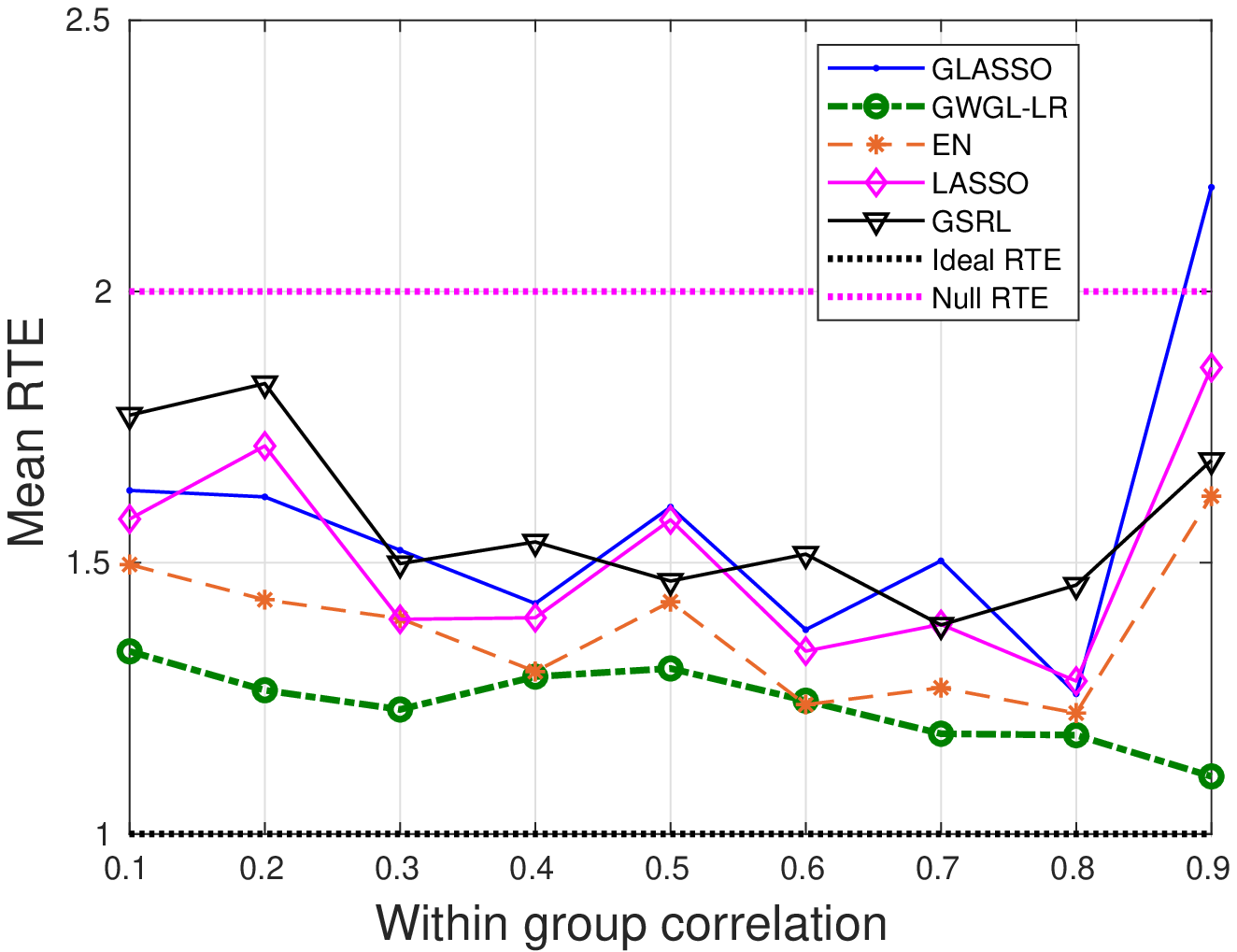}
		\caption{\small{Relative test error.}}
	\end{subfigure}%
	\begin{subfigure}{0.49\textwidth}
		\centering
		\includegraphics[width=1\textwidth]{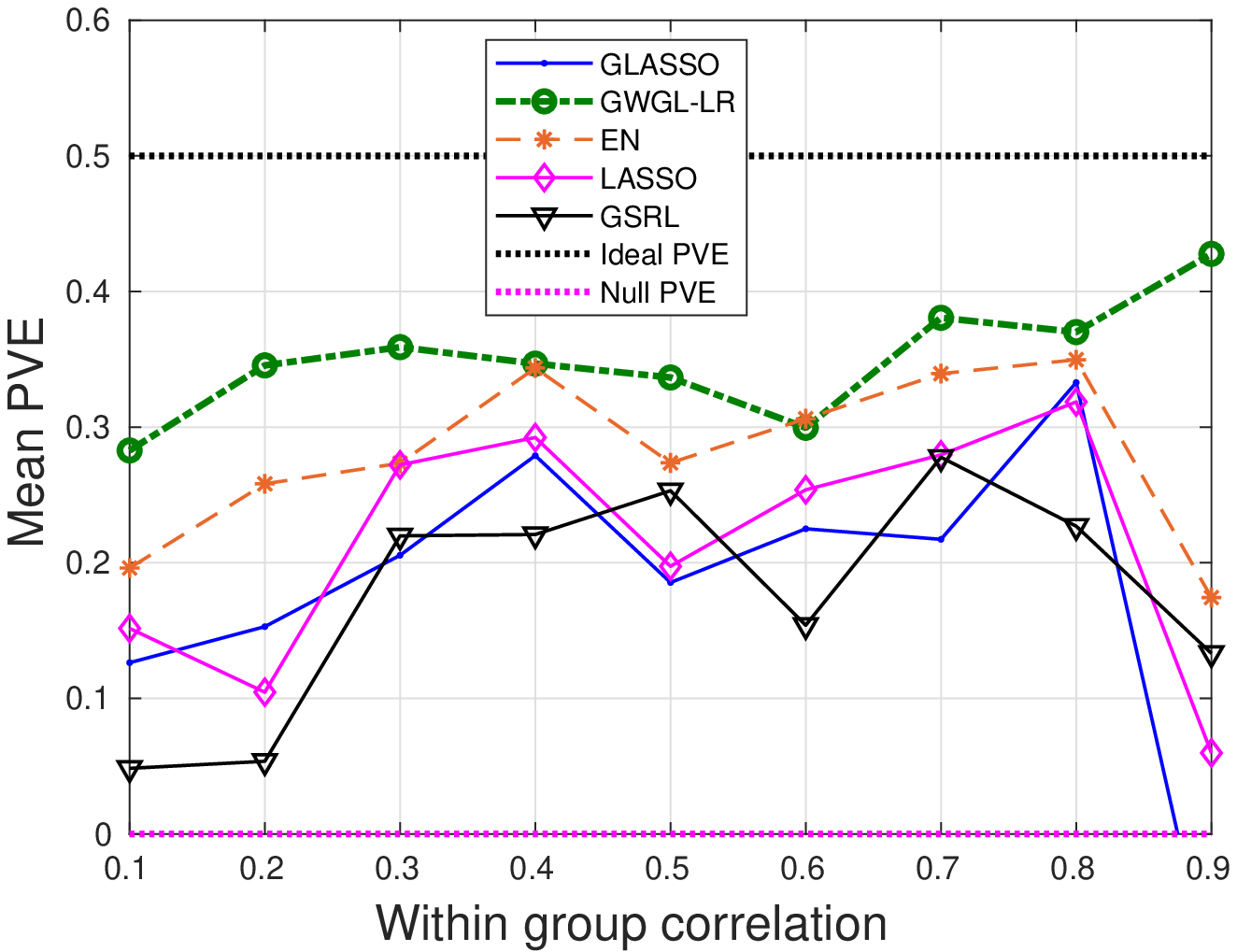}
		\caption{\small{Proportion of variance explained.}}
	\end{subfigure}
	\caption{The impact of within group correlation on the performance metrics, $q =
		20\%$.} \label{cor-20}
\end{figure}

See Figs. \ref{snr-20} and \ref{cor-20}.

\subsection*{Omitted Results in Section \ref{surgery}}
\subsubsection*{Pre-processing the Dataset}
Data were pre-processed as follows: (i) categorical variables (such as race,
discharge destination, insurance type) were numerically encoded and units
homogenized; (ii) missing values were replaced by the mode; (iii) all variables were
normalized by subtracting the mean and divided by the standard deviation; (iv)
patients who died within 30 days of discharge or had a postoperative length of stay
greater than 30 days were excluded.

\end{document}